%% file: neurips_2025.tex
\documentclass{article}

\usepackage[final]{neurips_2025}

\usepackage[utf8]{inputenc} 
\usepackage[T1]{fontenc}    
\usepackage{hyperref}       
\usepackage{url}            
\usepackage{booktabs}       
\usepackage{amsfonts}       
\usepackage{nicefrac}       
\usepackage{microtype}      
\usepackage{xcolor}         
\usepackage{array}
\usepackage{amsmath}
\usepackage{amsthm}
\usepackage{mathrsfs}
\usepackage{bm}

\usepackage{booktabs}       
\usepackage{makecell}      

\usepackage{multirow}
\usepackage{caption}
\usepackage{subfigure}
\usepackage{graphicx}
\usepackage{color}
\usepackage{amssymb}
\usepackage{enumitem}
\usepackage{wrapfig}

\usepackage{pifont}    
\usepackage{colortbl}
\usepackage{xcolor} 
\usepackage{float}

\newtheorem{theorem}{Theorem}[section] 
\newtheorem{proposition}[theorem]{Proposition} 

\title{GraphChain: Large Language Models for Large-scale Graph Analysis via Tool Chaining}

\author{%
    Chunyu Wei\textsuperscript{\rm 1}, Wenji Hu\textsuperscript{\rm 1}, Xingjia Hao\textsuperscript{\rm 2}, Xin Wang\textsuperscript{\rm 1}, Yifan Yang\textsuperscript{\rm 3}, Yueguo Chen\textsuperscript{\rm 1} \\ 
    \textbf{Yang Tian}\textsuperscript{\rm 2}, \textbf{Yunhai Wang}\textsuperscript{\rm 1}\thanks{Corresponding author.} \\
    \textsuperscript{\rm 1}Renmin University of China, China 
    \textsuperscript{\rm 2}Guangxi University, China \\
    \textsuperscript{\rm 3}Beijing Jiaotong University, China \\
    \texttt{weicy15@icloud.com}, 
    \texttt{2024000991@ruc.edu.cn}, 
    \texttt{haoxingjia@st.gxu.edu.cn} \\
    \texttt{2023103702@ruc.edu.cn},
    \texttt{23281027@bjtu.edu.cn} \\
    \texttt{chenyueguo@ruc.edu.cn}, 
    \texttt{ytian@gxu.edu.cn},
    \texttt{cloudseawang@gmail.com} \\
}

\begin{document}

\maketitle

\begin{abstract}
Large Language Models (LLMs) face significant limitations when applied to large-scale graphs, struggling with context constraints and inflexible reasoning. We present GraphChain, a framework that enables LLMs to analyze complex graphs through dynamic sequences of specialized tools, mimicking human exploratory intelligence. Our approach introduces two key innovations: (1) Progressive Graph Distillation, a reinforcement learning mechanism that generates optimized tool sequences balancing task relevance with information compression, and (2) Structure-aware Test-Time Adaptation, which efficiently tailors tool selection strategies to diverse graph topologies using spectral properties and lightweight adapters without costly retraining. Experiments show GraphChain significantly outperforms prior methods, enabling scalable and adaptive LLM-driven graph analysis. 
\footnote{ The code is available in \href{https://github.com/wuanjunruc/GraphChain}{https://github.com/wuanjunruc/GraphChain}}
\end{abstract}

\input{data/introduction}
\input{data/related_work}
\input{data/preliminary}
\input{data/methodology}
\input{data/experiment}

\input{data/conclusion}
\begin{ack}
This research was supported by NSFC (No.6250072448, No.62132017 and No.U2436209), the Shandong Provincial Natural Science Foundation (No.ZQ2022JQ32), the Beijing Natural Science Foundation (L247027), the Fundamental Research Funds for the Central Universities, the Research Funds of Renmin University of China, and the Young Elite Scientists Sponsorship Program by CAST under contract No. 2022QNRC001. It was also supported by Big Data and Responsible Artificial Intelligence for National Governance, Renmin University of China.
\end{ack}

\bibliographystyle{abbrvnat}
\bibliography{references}

\newpage

\section*{NeurIPS Paper Checklist}

\begin{enumerate}

\item {\bf Claims}
    \item[] Question: Do the main claims made in the abstract and introduction accurately reflect the paper's contributions and scope?
    \item[] Answer: \answerYes{} 
    \item[] Justification: Our abstract and introduction clearly claim our task (scope), contributions and solutions.

\item {\bf Limitations}
    \item[] Question: Does the paper discuss the limitations of the work performed by the authors?
    \item[] Answer: \answerYes{} 
    \item[] Justification: We discuss the limitation in Conclusion.

\item {\bf Theory assumptions and proofs}
    \item[] Question: For each theoretical result, does the paper provide the full set of assumptions and a complete (and correct) proof?
    \item[] Answer: \answerYes{} 
    \item[] Justification: We provide all the proof in Appendix~\ref{app:proof_ib_prop}.

    \item {\bf Experimental result reproducibility}
    \item[] Question: Does the paper fully disclose all the information needed to reproduce the main experimental results of the paper to the extent that it affects the main claims and/or conclusions of the paper (regardless of whether the code and data are provided or not)?
    \item[] Answer: \answerYes{} 
    \item[] Justification: We provide our code in the supplementary material. 

\item {\bf Open access to data and code}
    \item[] Question: Does the paper provide open access to the data and code, with sufficient instructions to faithfully reproduce the main experimental results, as described in supplemental material?
    \item[] Answer: \answerYes{} 
    \item[] Justification: We provide our code in the supplementary material. And we provide a github repository containing the code in https://github.com/GraphChain651/GraphChain.

\item {\bf Experimental setting/details}
    \item[] Question: Does the paper specify all the training and test details (e.g., data splits, hyperparameters, how they were chosen, type of optimizer, etc.) necessary to understand the results?
    \item[] Answer: \answerYes{} 
    \item[] Justification: In the Appendix~\ref{app:details_of_exp} and~\ref{app:data_construction}.

\item {\bf Experiment statistical significance}
    \item[] Question: Does the paper report error bars suitably and correctly defined or other appropriate information about the statistical significance of the experiments?
    \item[] Answer: \answerYes{} 
    \item[] Justification:  We conduct two-sample t-tests, and p-value < 0.05 indicates that the improvements are statistically significant.

\item {\bf Experiments compute resources}
    \item[] Question: For each experiment, does the paper provide sufficient information on the computer resources (type of compute workers, memory, time of execution) needed to reproduce the experiments?
    \item[] Answer: \answerYes{} 
    \item[] Justification: In the Appendix~\ref{app:details_of_exp}.
    
\item {\bf Code of ethics}
    \item[] Question: Does the research conducted in the paper conform, in every respect, with the NeurIPS Code of Ethics \url{https://neurips.cc/public/EthicsGuidelines}?
    \item[] Answer: \answerYes{} 
    \item[] Justification: This research conforms, in every respect, with the NeurIPS Code of Ethics.

\item {\bf Broader impacts}
    \item[] Question: Does the paper discuss both potential positive societal impacts and negative societal impacts of the work performed?
    \item[] Answer: \answerYes{} 
    \item[] Justification: In appendix~\ref{app:impact}. 
    
\item {\bf Safeguards}
    \item[] Question: Does the paper describe safeguards that have been put in place for responsible release of data or models that have a high risk for misuse (e.g., pretrained language models, image generators, or scraped datasets)?
    \item[] Answer: \answerNA{} 
    \item[] Justification: Not Applicable.

\item {\bf Licenses for existing assets}
    \item[] Question: Are the creators or original owners of assets (e.g., code, data, models), used in the paper, properly credited and are the license and terms of use explicitly mentioned and properly respected?
    \item[] Answer: \answerYes{} 
    \item[] Justification: We cite the original papers or website links about the dataset and open-source codes.

\item {\bf New assets}
    \item[] Question: Are new assets introduced in the paper well documented and is the documentation provided alongside the assets?
    \item[] Answer: \answerNA{} 
    \item[] Justification: Not Applicable.

\item {\bf Crowdsourcing and research with human subjects}
    \item[] Question: For crowdsourcing experiments and research with human subjects, does the paper include the full text of instructions given to participants and screenshots, if applicable, as well as details about compensation (if any)? 
    \item[] Answer: \answerNA{} 
    \item[] Justification: Not Applicable.

\item {\bf Institutional review board (IRB) approvals or equivalent for research with human subjects}
    \item[] Question: Does the paper describe potential risks incurred by study participants, whether such risks were disclosed to the subjects, and whether Institutional Review Board (IRB) approvals (or an equivalent approval/review based on the requirements of your country or institution) were obtained?
    \item[] Answer: \answerNA{} 
    \item[] Justification: Not Applicable.

\item {\bf Declaration of LLM usage}
    \item[] Question: Does the paper describe the usage of LLMs if it is an important, original, or non-standard component of the core methods in this research? Note that if the LLM is used only for writing, editing, or formatting purposes and does not impact the core methodology, scientific rigorousness, or originality of the research, declaration is not required.
    \item[] Answer: \answerNA{} 
    \item[] Justification:  Not Applicable.

\end{enumerate}

\newpage
\appendix

\input{app/proof_of_IB}
\input{app/case_study}
\input{app/experiment_setup}
\input{app/baseline_app}
\input{app/nx_function}
\input{app/dataset}

\input{app/stta_complexity}
\input{app/impact}


\end{document}

%% file: data/introduction.tex
\section{Introduction}
\label{sec:intro}

Graph-structured data represents a fundamental paradigm across diverse domains, from social networks and molecular structures to knowledge bases and recommendation systems. While large language models (LLMs) have demonstrated remarkable reasoning capabilities, they encounter significant challenges when processing graph data.

\begin{figure}[!t]
  \centering
  \includegraphics[width=0.99\textwidth]{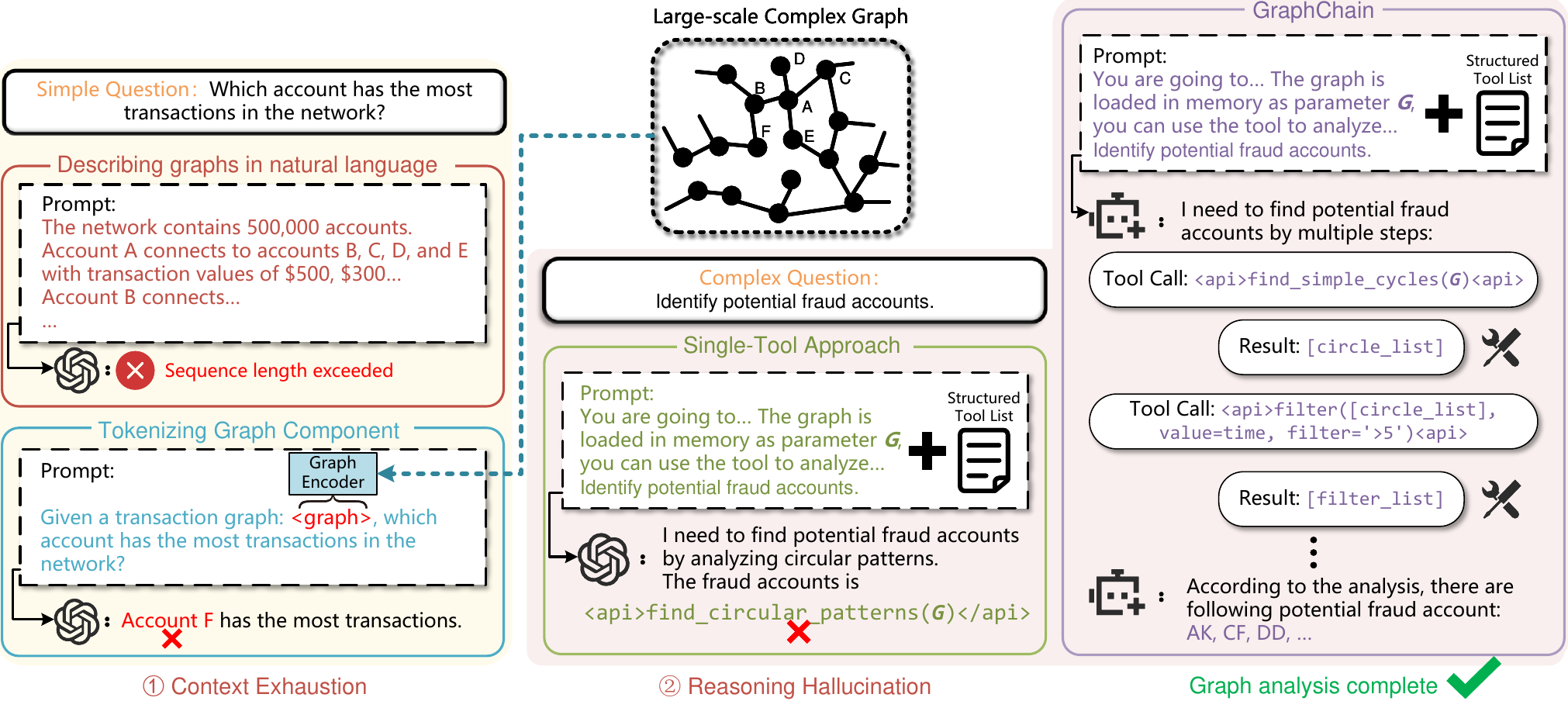}
  \caption{\small \textbf{Comparison of Graph Processing Approaches with LLMs.} \underline{Left}: Methods suffer from Context Exhaustion where large graphs exceed LLM context windows. \underline{Center}: Single-tool approaches face Reasoning Hallucination with fixed, predefined tools. \underline{Right}: Our \texttt{GraphChain} framework enables human-like exploratory analysis through sequential tools that progressively narrow focus in large-scale graphs.}
  \label{fig:illustration}
\end{figure}

Recent approaches to enhancing LLMs' graph processing capabilities have taken two primary directions. The first attempts to adapt LLMs to directly process graph structures—either through tokenization or natural language descriptions~\citep{chai2023graphllm, wang2023can}. However, this approach faces \textbf{Context Exhaustion}: large-scale graphs with millions of nodes and edges cannot be effectively compressed within LLMs' context limitations, making it computationally infeasible to load entire subgraphs into their context windows (Figure~\ref{fig:illustration}, left).

Recognizing these limitations, a second direction draws inspiration from tool learning paradigms. Approaches like Graph-ToolFormer~\citep{DBLP:journals/corr/abs-2304-11116} and GraphForge~\citep{wang2024graphtool} pioneered integrating specialized tools with LLMs for graph reasoning, enabling models to call external graph processing functions. However, these methods primarily conceptualize tool learning as text generation, relying on single-step tool invocations with textually described graph structures. This approach leads to \textbf{Reasoning Hallucination} (Figure~\ref{fig:illustration}, middle), as it places unrealistic demands on individual tools to provide comprehensive functionality for complex graph analysis.

Complex graph analysis parallels human exploration of unknown environments. Just as humans navigate unfamiliar territories through interactive, adaptive exploration—where each step reveals information that guides subsequent decisions—effective graph analysis requires progressive, sequential information gathering rather than comprehensive analysis in one operation. A field researcher might first survey an area broadly before focusing on regions of interest; similarly, graph analysis benefits from incremental understanding built through sequential operations.

Inspired by human exploratory cognition, we propose \texttt{GraphChain}, a novel framework enabling LLMs to process large-scale graphs through dynamic tool-chaining (Figure~\ref{fig:illustration}, right). \texttt{GraphChain} decomposes complex graph problems into sequences of specialized operations, activating LLMs' reasoning capabilities to create, refine, and execute chains of graph processing tools. This approach allows progressive refinement and deeper exploration of graph structures, mimicking how human experts methodically investigate complex systems layer by layer.

The implementation of \texttt{GraphChain} addresses two significant technical challenges:
\begin{enumerate}[leftmargin=*]
    \item \textbf{Informative Tool Sequence Generation} requires determining optimal tool sequences for diverse analytical tasks, navigating an exponentially growing space of possible combinations. Traditional approaches struggle with this challenge due to scarce labeled data for complex graph analysis.
    
    \item \textbf{Adaptive Graph Structure Sensing} must address real-world graph data exhibiting distributional shifts and structural variations. Unlike natural data types, graph structures are heavily human-defined with domain-specific schemas, leading to severe distribution shifts across domains. 
\end{enumerate}

To generate informative tool sequences, we propose a progressive graph distillation training mechanism. Our key insight is that effective graph analysis mirrors human exploration: beginning broadly and systematically narrowing focus as relevant information emerges. This approach transforms the exponential tool-selection problem into a principled information bottleneck optimization, iteratively refining both structural scope and representational complexity while preserving only task-critical information—similar to how humans selectively attend to relevant environmental cues.

For adapting to diverse graphs, we introduce a structure-aware test-time adaptation mechanism. We leverage the insight that graph topology fundamentally influences optimal analysis strategies, just as explorers adjust techniques for different terrains. Our lightweight adapter dynamically modifies tool selection policy based on spectral properties capturing essential structural characteristics, enabling \texttt{GraphChain} to maintain effectiveness across diverse graphs while preserving efficiency.

Our main contributions include:
\begin{itemize}[leftmargin=*]
    \item \texttt{GraphChain}, a novel framework leveraging Graph-Oriented Reinforcement Learning with progressive information distillation, enabling systematic exploration of large-scale graphs through interconnected tool sequences.
    
    \item A structure-aware test-time adaptation mechanism that adjusts tool-chaining strategies based on graph topology, enabling efficient transfer to diverse graph structures without costly retraining.

    \item Extensive experimentation demonstrating that \texttt{GraphChain} significantly outperforms existing methods by an average of 20.7\%, with exceptional scalability handling graphs up to 200,000 nodes while maintaining consistent performance.
\end{itemize}

%% file: data/related_work.tex
\section{Related Work}
\label{sec:related}

\paragraph{Tool Learning for LLMs}
Tool learning for LLMs encompasses tuning-free methods using prompting strategies like Chain-of-Thought~\citep{wei2022chain}, ReAct~\citep{yao2023react}, and DFSDT~\citep{qin2023toolllm}, alongside approaches integrating tools into conversations~\citep{chen2023chatcot} or employing structured selection via graphs~\citep{liu2024toolnet}, hierarchies~\citep{du2024anytool}, or intent filtering~\citep{fore2024geckopt}. Meanwhile, tuning-based methods directly adapt LLM parameters~\citep{xu2023tool} through behavior cloning with reinforcement learning~\citep{qiao2024making, yu2024steptool}, fine-tuning on specialized decision data~\citep{qin2023toolllm}, frameworks for varying tool complexities~\citep{gao2024confucius}, and self-verification mechanisms~\citep{mekala2024toolverifier}.

\paragraph{Graph Processing with LLMs}
Recent work enhances LLMs for graph processing via: (1) Direct processing with text or visual graph descriptions~\citep{wang2023can, guo2023gpt4graph} or specialized token sequences~\citep{chen2024llaga, wang2024instructgraph}; (2) Tool integration and agent-based methods for external function calls~\citep{zhang2023graph} or multi-step reasoning~\citep{gu2024middleware}; (3) GNN-LLM combinations using GNNs as encoders~\citep{tang2024graphgpt} or aligning representation spaces~\citep{su2022molecular}.

\paragraph{Test-time Adaptation}

Traditional machine learning assumes identical training and testing distributions, but real-world deployments often encounter distribution shifts \citep{kulinski2023towards}. Test-Time Adaptation (TTA) addresses this challenge \citep{liang2025comprehensive, alfarra2025test}. For LLMs, adaptation techniques include test-time prompt tuning \citep{shu2022test, ma2023swapprompt}, Parameter-Efficient Fine-Tuning methods like adapters or LoRA \citep{hu2022lora} for efficient updates \citep{shi2024medadapter, muhtar2024streamadapter}, and "test-time compute scaling" with iterative refinement, search, or self-correction \citep{jaech2024openai, guo2025deepseek, suzgun2025dynamic}.

%% file: data/preliminary.tex
\section{Preliminaries and Problem Formulation}
\label{sec:preliminaries}

\paragraph{Graph Notation}

Let $G = (\mathcal{V}, \mathcal{E})$ represent a graph, where $\mathcal{V} = \{v_1, v_2, \ldots, v_n\}$ is the set of $n=|\mathcal{V}|$ nodes and $\mathcal{E} \subseteq \mathcal{V} \times \mathcal{V}$ is the set of $m=|\mathcal{E}|$ edges. The adjacency matrix $\mathbf{A} \in \{0, 1\}^{n \times n}$ (or $\mathbb{R}^{n \times n}$ for weighted graphs) has entries $\mathbf{A}_{ij} = 1$ (or edge weight) if $(v_i, v_j) \in \mathcal{E}$, and 0 otherwise. Node features are represented by matrix $\mathbf{X} \in \mathbb{R}^{n \times d}$. The degree matrix $\mathbf{D}$ is diagonal with $\mathbf{D}_{ii} = \sum_{j=1}^n \mathbf{A}_{ij}$, and the normalized graph Laplacian is defined as $\mathbf{L} = \mathbf{I} - \mathbf{D}^{-1/2}\mathbf{A}\mathbf{D}^{-1/2}$. A node's neighborhood is $\mathcal{N}(v) = \{u \in \mathcal{V} \mid (v, u) \in \mathcal{E}\}$, and a subgraph $G' = (\mathcal{V}', \mathcal{E}')$ consists of node subset $\mathcal{V}' \subseteq \mathcal{V}$ and edge subset $\mathcal{E}' \subseteq \mathcal{E} \cap (\mathcal{V}' \times \mathcal{V}')$.

\paragraph{Graph Processing Tool Library}


We define a library of graph processing tools $\mathcal{T} = \{T_1, T_2, \ldots, T_K\}$, which are implemented based on functions from the NetworkX library\footnote{https://networkx.org.}, and operate on tensor representations within the current \textbf{memory state} $\mathbf{m}$. NetworkX is a widely-used open-source Python library that provides a comprehensive set of graph processing functions, including node and edge operations, graph property calculations, and advanced analytical tools. The tools in the library are based on 45 carefully selected NetworkX functions, with details provided in Appendix~\ref{app:tool_library}. The tools operate on tensor representations, typically containing the adjacency matrix $\mathbf{A}'$ and feature matrix $\mathbf{X}'$ for a subgraph $G' = (\mathcal{V}', \mathcal{E}')$:

\begin{equation}
\mathbf{m} \approx (\mathbf{A}' \in \mathbb{R}^{n' \times n'}, \mathbf{X}' \in \mathbb{R}^{n' \times d}, \dots) \quad \text{where } n' = |\mathcal{V}'|
\end{equation}

A tool $T$ takes the current memory state $\mathbf{m}$ and tool-specific parameters $\theta_T$ as input, producing two outputs: (1) A concise natural language summary $d$ of the execution outcome; (2) An updated memory state $\mathbf{m}'$. 
Formally, the tool function is defined as: \quad
$T: (\mathbf{m}, \theta_T) \mapsto (d, \mathbf{m}')$. 

This dual output mechanism allows our framework to provide context-window-friendly summaries to the LLM via $d$, while managing potentially large-scale intermediate graph data within $\mathbf{m}'$, mitigating context exhaustion when processing large graphs.

\paragraph{Sequential Graph Exploration as an MDP}
Given an analytical query $\mathcal{Q}$ and input graph $G$, we model sequential graph exploration as a Markov Decision Process (MDP) $M = (\mathcal{S}, \mathcal{A}, P, R, \gamma)$:
\begin{itemize}[leftmargin=*]
    \item \textbf{State Space} $\mathcal{S}$: State $s_t$ encapsulates query $\mathcal{Q}$, graph reference, action history $\{(a_i, d_i)\}_{i=0}^{t-1}$, and memory state $\mathbf{m}_{t-1}$.
    
    \item \textbf{Action Space} $\mathcal{A}$: Actions $a_t = (T, \theta_T)$ select a tool $T \in \mathcal{T}$ with parameters $\theta_T$, or 'TERMINATE'.
    
    \item \textbf{Transition Dynamics} $P$: Tool execution produces $(d_t, \mathbf{m}_t) = T(\mathbf{m}_{t-1}, \theta_T)$, updating state $s_{t+1}$ with new history and memory.
    
    \item \textbf{Reward Function} $R(s_t, a_t, s_{t+1})$: Evaluates actions based on progress and task success.
    
    \item \textbf{Discount Factor} $\gamma \in [0, 1]$: Balances immediate vs. future rewards.
\end{itemize}
The agent's policy $\pi_\theta(a_t|s_t)$, parameterized by $\theta$, generates a trajectory $\tau = \{s_1, a_1, s_2, a_2, ..., s_T, a_T\}$ representing sequential tool interactions. To maximize performance, we optimize the expected reward:
{\small
\begin{equation}
\nabla \overline{R_\theta} = \sum_{\tau} R(\tau) \nabla \pi_\theta(\tau) = \mathbb{E}_{\tau \sim \pi_\theta, (s_t,a_t) \sim \tau} \left[R(\tau) \sum_{t=1}^{T} \nabla_\theta \log \pi_\theta(a_t|s_t)\right]
\end{equation}
}

%% file: data/methodology.tex
\section{Methodology}
\label{sec:methodology}
\begin{figure}[!t]
  \centering
  \includegraphics[width=0.99\textwidth]{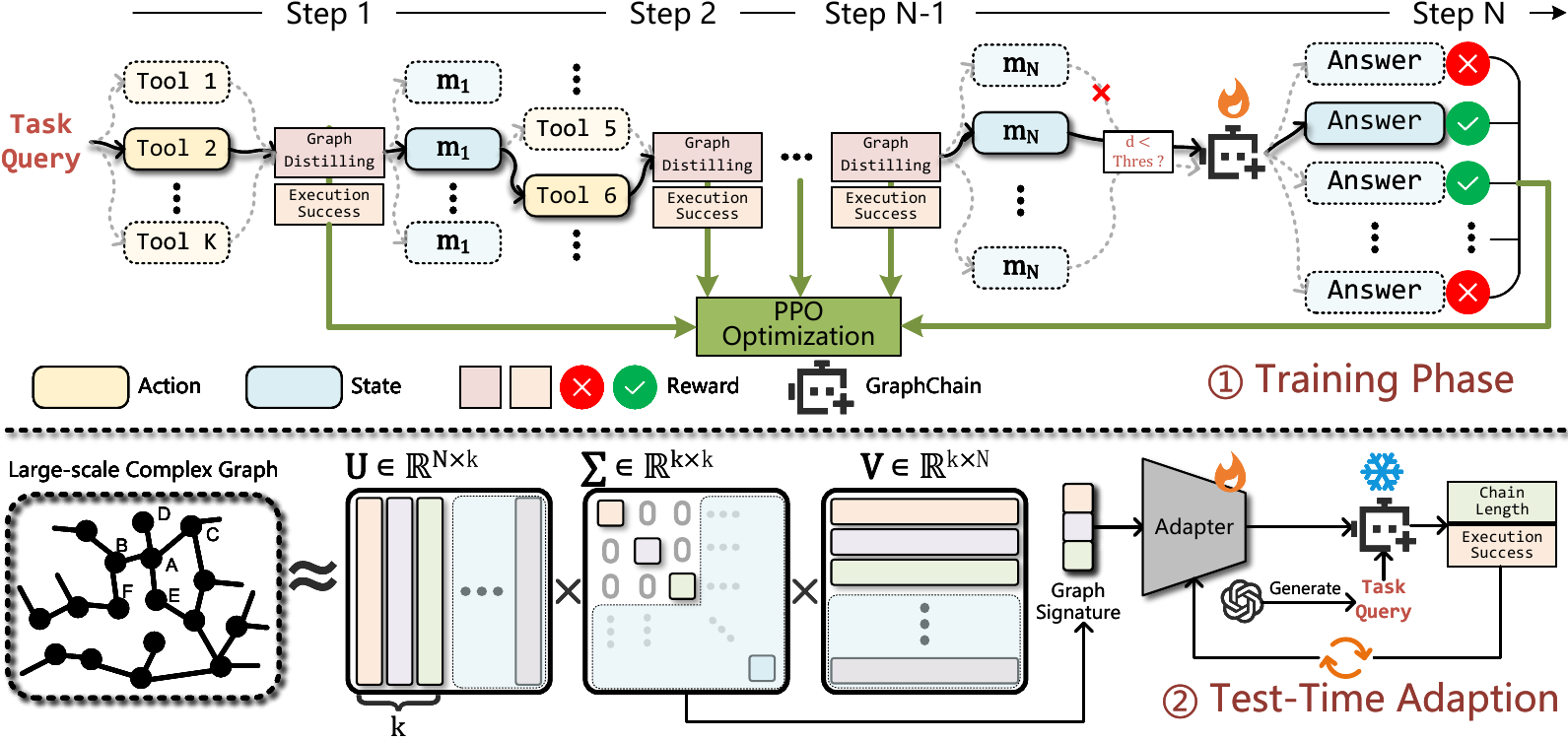}
  \caption{\small \textbf{(1) Training Phase:} Progressive graph distillation where the RL agent learns to select tool sequences that iteratively reduce the memory state's ($\mathbf{m}$) Graph Description Length (GDL) while maximizing task relevance. \textbf{(2) Structure-aware Test-Time Adaptation:} A lightweight adapter ($\mathcal{A}_{\psi}$) tuned by minimizing chain length and KL divergence generates a structure-specific soft prompt $\mathbf{P}_G$ based on the graph's SVD-derived fingerprint $\mathbf{z}_G$.}
  \label{fig:framework}
\end{figure}
GraphChain addresses the challenges of applying LLMs to large-scale graph analysis by formulating the problem as a sequential decision-making task solvable via reinforcement learning. Our approach centers on two core technical innovations: (1) \textbf{Progressive Graph Distillation}, which promotes informative yet compact state representations, and (2) \textbf{Structure-aware Test-Time Adaptation}, enabling dynamic adjustment to diverse graph topologies. Figure~\ref{fig:framework} provides a conceptual overview.

\subsection{Progressive Graph Distillation}
\label{sec:distillation_rl}

Generating effective tool sequences for complex graph queries involves navigating an exponentially large action space. To provide denser learning signals and emulate human-like analytical workflows that progress from coarse to fine, we introduce Progressive Graph Distillation.

This approach incentivizes the RL agent to pursue both the query objective and manage the complexity of its memory state $\mathbf{m}$. We train the agent to prioritize tool sequences that systematically reduce $\mathbf{m}$'s data volume while retaining task-critical information, transforming exploration into a guided search characterized by iterative refinement. The aim is to progressively shrink $\mathbf{m}$ step-by-step, eventually yielding a compact final state $\mathbf{m}_N$ suitable for direct processing within the LLM's context window.

\subsubsection{Quantifying Memory State Volume and Relevance}
\label{sec:volume_relevance}

Implementing progressive distillation requires quantifying two key aspects of memory state $\mathbf{m}_t$ at each step $t$: its \textbf{data volume} and its \textbf{relevance} to query $\mathcal{Q}$.

\paragraph{Graph Description Length ($\mathrm{GDL}(\mathbf{m}_t)$):} Drawing from the Minimum Description Length principle, we introduce Graph Description Length to measure the data size needed to represent the current graph state. Assuming memory state $\mathbf{m}_t$ contains subgraph $G'_t = (\mathcal{V}'_t, \mathcal{E}'_t)$ with $n'_t = |\mathcal{V}'_t|$ nodes and $m'_t = |\mathcal{E}'_t|$ edges, plus node features $\mathbf{X}'_t \in \mathbb{R}^{n'_t \times d_f}$, we define:
\begin{equation}
\mathrm{GDL}(\mathbf{m}_t) = L(\text{structure}) + L(\text{features}) \approx \alpha_s m'_t + \alpha_f n'_t d_f
\label{eq:memory_gdl}
\end{equation}
Coefficients $\alpha_s, \alpha_f \ge 0$ weight the relative contribution of structural versus feature information.
\vspace{-1em}
\paragraph{Task Relevance ($\mathrm{Rel}(\mathbf{m}_t, \mathcal{Q})$):} We employ an auxiliary LLM scorer to assess the utility of $\mathbf{m}_t$ for answering query $\mathcal{Q}$. Since $\mathbf{m}_t$ may exceed the LLM's context limits, we use the concise description $d_t$ produced by the executed tool. We estimate the task relevance by:
\begin{equation}
\mathrm{Rel}(\mathbf{m}_t, \mathcal{Q}) \approx \text{LLMScore}(\text{prompt}(\mathcal{Q}, H_t, d_t)) \in [0, 1]
\label{eq:task_relevance}
\end{equation}
where $H_t = \{d_0, \dots, d_{t-1}\}$ is the history of preceding descriptions.

\subsubsection{Distillation-based Reward Shaping}
\label{sec:reward_design}

We incorporate progressive distillation into the RL reward function $R_t = R(s_t, a_t, s_{t+1})$. The reward structure provides feedback during exploration while assessing final task completion:
{\small
\begin{equation}
R_t =
\begin{cases}
w_1 \cdot \hat{r}^{\mathrm{Succ}}_t + w_2 \cdot \hat{r}^{\Delta \mathrm{GDL}}_t + w_3 \cdot \hat{r}^{\Delta \mathrm{Rel}}_t & \text{if } t < N \\
w_{\text{solve}} \cdot \mathrm{EvaluateTaskSuccess}(\mathcal{Q}, s_{N+1}) & \text{if } t = N
\end{cases}
\label{eq:unified_reward_distill}
\end{equation}
}
where $N$ is the final step index, and the intermediate reward components are:

\begin{itemize}[leftmargin=*,itemsep=1pt,topsep=2pt]
   \item $\hat{r}^{\mathrm{Succ}}_t = \mathbb{I}(\textit{ExecutionSuccess}(a_t, s_{t+1}))$: Binary reward for valid tool execution.

   \item $\hat{r}^{\Delta \mathrm{GDL}}_t = \tanh\left(\beta \frac{\mathrm{GDL}(\mathbf{m}_{t-1}) - \mathrm{GDL}(\mathbf{m}_t)}{\mathrm{GDL}(\mathbf{m}_{t-1}) + \epsilon}\right) \in (-1, 1)$: Rewards reduction in relative GDL. 

   \item $\hat{r}^{\Delta \mathrm{Rel}}_t = \text{Rel}_t - \text{Rel}_{t-1}$: Rewards increase in estimated task relevance.
\end{itemize}

Weights $w_1, w_2, w_3$ balance the importance of execution success, volume reduction, and relevance gain. Weight $w_{\text{solve}}$ scales the final reward based on overall success in addressing query $\mathcal{Q}$.

\subsubsection{Information Bottleneck Perspective}
\label{sec:info_bottleneck}

Our progressive distillation mechanism aligns with the Information Bottleneck principle, advocating for representations that are maximally informative about a target while being maximally compressive of input. Our reward function operationalizes this trade-off by incentivizing high task relevance while rewarding reductions in state volume.

\begin{proposition}
\label{prop:ib_compression}
Let the input be $X=(G, \mathcal{Q})$, containing task-relevant information $Y=\mathcal{A}_{\mathcal{Q}}$ (the answer) and task-irrelevant information $IR$, with the Markov structure $(Y, IR) \rightarrow X \rightarrow \mathbf{m}_t$. Assuming the relevance proxy $\mathrm{Rel}_t$ positively correlates with the mutual information $I(\mathbf{m}_t; Y)$ and the GDL serves as a complexity measure encouraging smaller $I(X; \mathbf{m}_t)$, optimizing policy $\pi_\theta$ with reward function $R_t$ guides the generation of memory states $\mathbf{m}_t$ that tend to minimize irrelevant information $I(IR; \mathbf{m}_t | Y)$ while preserving relevant information $I(\mathbf{m}_t; Y)$.
\end{proposition}

Detailed proof is provided in Appendix~\ref{app:proof_ib_prop}. This proposition provides theoretical support for our distillation approach. By rewarding both relevance gain and volume reduction, the RL process steers the agent toward behaviors that effectively filter graph data—reducing the representational footprint of task-irrelevant components while preserving critical information.

\subsubsection{Policy Optimization}
\label{sec:policy_opt}

To optimize the LLM agent's policy $\pi_\theta$, we implement Proximal Policy Optimization (PPO), using Generalized Advantage Estimation (GAE) for improved stability:
{\small
\begin{equation}
\hat{A}_t^{\text{GAE}}(\theta, \omega) = \sum_{l=0}^{N-t} (\gamma\lambda)^l \delta_{t+l}, \quad \text{where} \quad \delta_t = R_{t+1} + \gamma V_\omega(s_{t+1}) - V_\omega(s_t)
\label{eq:gae_distill_revised}
\end{equation}
}
Here, $\lambda \in [0, 1]$ is the GAE trace decay parameter, $V_\omega$ is the learned value function, $\gamma$ is the discount factor, and $R_{t+1}$ is the distillation-aware reward.

Following the PPO-clip approach, we maximize a clipped surrogate objective based on trajectories $\tau$ sampled from policy $\pi_\theta$:
{\small
\begin{equation}
\mathcal{L}^{\text{CLIP}}(\theta) = \hat{\mathbb{E}}_{\tau \sim \pi_\theta} \left[ \sum_{t=0}^{N} \min \left( \frac{\pi_\theta(a_t|s_t)}{\pi_{\theta_{\text{old}}}(a_t|s_t)} \hat{A}_t^{\text{GAE}}, \, \text{clip}\left(\frac{\pi_\theta(a_t|s_t)}{\pi_{\theta_{\text{old}}}(a_t|s_t)}, 1 - \epsilon, 1 + \epsilon \right) \hat{A}_t^{\text{GAE}} \right) \right]
\label{eq:ppo_clip_obj_revised}
\end{equation}
}
where $\pi_{\theta_{\text{old}}}$ is the old policy used for generating trajectories, and $\epsilon$ is the clipping hyperparameter.

\subsection{Structure-aware Test-Time Adaptation}
\label{sec:stta_spt_revised}


\subsubsection{Graph Structural Fingerprinting}
\label{sec:fingerprint_svd_revised}

To provide global structural awareness for large-scale graphs, we derive a concise graph fingerprint. We compute the normalized graph Laplacian $\mathbf{L} = \mathbf{I} - \mathbf{D}^{-1/2}\mathbf{A}\mathbf{D}^{-1/2}$ and consider its Singular Value Decomposition, $\mathbf{L} = \mathbf{U} \mathbf{\Sigma} \mathbf{V}^T$. The smallest singular values $\sigma_i$ capture dominant, low-frequency components reflecting macroscopic graph properties. We define the \textbf{structural fingerprint} as:
$\mathbf{z}_G = (\sigma_0, \sigma_1, \ldots, \sigma_M) \in \mathbb{R}^{M+1}$. 

While full SVD is intractable for very large graphs, these $M+1$ smallest singular values (where $M \ll N$) can be computed efficiently using iterative algorithms, effectively distilling essential global topology into a compact vector. We provide complexity analysis in Appendix~\ref{app:stta_complexity}.

\subsubsection{Structure-Conditioned Prompt Generation}
\label{sec:spt_adapter}

STTA employs a continuous adaptation mechanism through adapter network $\mathcal{A}_{\psi}$, which maps the graph's structural fingerprint $\mathbf{z}_G$ to a soft prompt $\mathbf{P}_G = \mathcal{A}_{\psi}(\mathbf{z}_G) \in \mathbb{R}^{L_p \times d_{emb}}$:

This generated prompt is prepended to the standard embedding $E(s_t)$ of the agent's state, modifying the input to the frozen LLM policy:~~$\text{LLMInput}(s_t, G) = [\mathbf{P}_G ; E(s_t)] = [\mathcal{A}_{\psi}(\mathbf{z}_G) ; E(s_t)]$. 

The agent's action is then sampled from the policy conditioned on this augmented input. During adaptation, only the smaller set of adapter parameters $\psi$ are tuned, enabling efficient adaptation.

\subsubsection{Self-Supervised Adaptation}
\label{sec:tta_self_supervised}

Given the absence of ground-truth rewards for user query $\mathcal{Q}$ on unseen test graph $G_{\text{test}}$, STTA employs a self-supervised strategy using auxiliary queries. We leverage a general-purpose LLM to generate diverse auxiliary graph analysis queries relevant to $G_{\text{test}}$'s structure.

For each auxiliary query, we perform rollouts using the frozen base policy conditioned on the graph-specific prompt, yielding trajectories. The adaptation objective balances planning efficiency and policy regularization:
{\small
\begin{equation}
L_{\text{STTA}}(\psi) = \mathbb{E}_{\mathcal{Q}_{\text{aux}, i}, \tau_i \sim \pi_{\psi}(\cdot|s; G_{\text{test}})} \left[ w_L N_{\tau_i} + w_{KL} \sum_{t=0}^{N_{\tau_i}-1} D_{KL}(\pi_{\psi}(\cdot | s_t; G_{\text{test}}) || \pi_{\text{orig}}(\cdot | s_t)) \right]
\label{eq:stta_kl_objective}
\end{equation}
}
The components of this objective are: (1) \textbf{Chain Length ($N_{\tau_i}$)} encouraging efficient planning, and (2) \textbf{KL Divergence Regularization} ensuring helpful but not drastic changes.

We minimize this objective using the REINFORCE algorithm, tuning $\mathcal{A}_{\psi}$ to generate prompts that enhance efficiency while maintaining fidelity to learned behaviors, effectively adapting the frozen policy to $G_{\text{test}}$'s specific structure.

%% file: data/experiment.tex
\section{Experiment}
\label{sec:experiment}
\subsection{Experimental Setting}
\textbf{Graph Dataset.} We evaluate \texttt{GraphChain} on five diverse graph datasets representing different real-world domains, as illustrated in Table~\ref{tab:dataset_stats}.

\begin{table}[h]
\centering
\caption{\small Statistics of graph datasets used in our experiments.}
\label{tab:dataset_stats}
\resizebox{\textwidth}{!}{%
\begin{tabular}{@{}lcccclp{4.5cm}@{}}
\toprule
\textbf{Scenario} & \textbf{Dataset} & \textbf{\#Nodes} & \textbf{\#Edges} & \textbf{\#Features} & \textbf{Type} & \textbf{Description} \\
\midrule
\multirow{3}{*}{Citation Graphs} & Cora & 2,708 & 10,556 & 1,433 & \multirow{3}{*}{Directed} & \multirow{3}{4.5cm}{Academic papers connected by citation relationships \citep{yang2016revisiting}} \\
 & CiteSeer & 3,327 & 9,104 & 3,703 & & \\
 & PubMed & 19,717 & 88,648 & 500 & & \\
\midrule
\multirow{2}{*}{Social Networks} & Facebook & 4,039 & 88,234 & - & Undirected & \multirow{2}{4.5cm}{Online interactions ~\citep{leskovec2012learning}} \\
 & Twitter & 81,306 & 1,768,149 & - & Directed & \\
\midrule
Chemical Molecules & QM9 & $\sim$18.0/graph & $\sim$37.3/graph & 11 & Undirected & Molecular structures with bonds between atoms \citep{wu2018moleculenet} \\
\midrule
Traffic Networks & METR-LA & 207 & 1,515 & - & Directed & Road networks with geographic constraints \citep{chen2020multi} \\
\midrule
Financial Networks & Elliptic & 203,769 & 234,355 & 165 & Directed & Transaction networks~\citep{weber2019anti} \\
\bottomrule
\end{tabular}%
}
\end{table}

\textbf{Instruction Data.}
We constructed two complementary datasets: (1) an SFT dataset comprising 9,986 (query, tool sequence, answer) triplets based on 45 carefully selected NetworkX functions, and (2) an RL dataset containing 3,000 expert-annotated (query, answer) pairs (600 per graph scenario). We allocated 500 pairs per scenario for training and 100 for testing, with domain experts crafting exemplary instruction templates to ensure ecological validity. See Appendix~\ref{app:data_construction} for details.

\textbf{Baselines.}
We evaluated \texttt{GraphChain} against state-of-the-art methods from two categories:

(1). For Text-Instruction methods, we tested leading closed-source LLMs (\texttt{Claude-series}~\citep{TheC3}, \texttt{GPT-series}~\citep{DBLP:journals/corr/abs-2303-08774}, and \texttt{GLM4-0520}~\citep{glm2024chatglm}) using two-shot prompting with Chain-of-Thought reasoning, and reproduced specialized graph reasoning methods (\texttt{NLGraph}~\citep{DBLP:conf/nips/WangFHTHT23}, \texttt{GraphWiz}~\citep{DBLP:journals/corr/abs-2402-16029}).

(2). For Tool-Instruction methods, we compared against recent tool-augmented approaches (\texttt{Graph-ToolFormer}~\citep{DBLP:journals/corr/abs-2304-11116}, \texttt{GraphForge}~\citep{wang2024graphtool}, and \texttt{ToolGen}~\citep{DBLP:conf/iclr/WangHJWB025}).

To ensure fair comparisons with existing baselines—all requiring the entire graph in the context window—we partitioned original graphs into subgraphs with fewer than 100 nodes for overall comparison. 
We use the same input for both baseline methods and \texttt{GraphForge}.
In our scalability experiment (Section~\ref{subsec:scalibility}), \texttt{GraphChain} maintains comparable performance even when scaling to graphs with approximately 200,000 nodes.
Further details are provided in Appendix~\ref{app:baseline}.

\textbf{Training Setup.}
We used two NVIDIA A800 GPUs with LoRA-based fine-tuning (rank $r$=16, alpha=32) on the \texttt{Qwen2.5-7B-instruction} model. 
Further details are provided in Appendix~\ref{app:details_of_exp}.

\subsection{Main Results}
\begin{table}[t]
\centering
\caption{\small Performance comparison (accuracy \%) across five real-world graph reasoning scenarios.}
\label{tab:main_results}
\resizebox{\textwidth}{!}{%
\begin{tabular}{@{}lccccccc@{}}
\toprule
\multicolumn{8}{c}{\textbf{Text-Instruction Methods}} \\
\midrule
\textbf{Model} & \textbf{Parameters} & \textbf{Financial Network} & \textbf{Chemical Molecule} & \textbf{Social Network} & \textbf{Citation Graph} & \textbf{Traffic Network} & \textbf{Average} \\
\midrule
\texttt{Claude-3-Sonnet} & - & 21.7 $\pm$ 1.8& 47.0 $\pm$ 2.2 & 21.5 $\pm$ 3.2& 17.7 $\pm$ 2.1& 16.8 $\pm$ 2.0& 24.9 $\pm$ 2.3 \\
\texttt{GPT-3.5-turbo} & $\sim$175B & 36.6 $\pm$ 2.1& 23.0 $\pm$ 3.7& 18.2 $\pm$ 3.6& 12.2 $\pm$ 0.8& 19.4 $\pm$ 1.9& 21.9 $\pm$ 2.4\\
\texttt{Claude-3-Haiku} & $\sim$20B & 12.2 $\pm$ 3.0& 52.9 $\pm$ 3.2& 50.3 $\pm$ 3.4& 19.8 $\pm$ 2.0& 13.9 $\pm$ 2.4& 29.8 $\pm$ 2.8\\
\texttt{Claude-3-Opus} & $\sim$137B & 23.6 $\pm$ 2.1& 42.4 $\pm$ 1.4& 51.9 $\pm$ 1.3& 36.7$\pm$ 3.1& 43.4 $\pm$ 3.3& 39.6 $\pm$ 2.2\\
\texttt{GraphWiz} & 13B & 41.1 $\pm$ 3.9& 52.4 $\pm$ 2.6& 61.5 $\pm$ 3.5& 68.0 $\pm$ 2.1& 38.4 $\pm$ 1.9& 52.3 $\pm$ 2.9\\
\texttt{NLGraph} & $\sim$100B & 52.1 $\pm$ 3.4& 58.4 $\pm$ 2.5& 65.2 $\pm$ 2.3& 59.4 $\pm$ 0.5& 39.8 $\pm$ 1.8& 55.0 $\pm$ 2.1\\
\texttt{GPT-4o} & $\sim$200B & 57.5 $\pm$ 1.9& 62.7 $\pm$ 3.6& 65.2 $\pm$ 3.9& 71.5 $\pm$ 3.4& 43.4 $\pm$ 1.6 & 59.4 $\pm$ 2.6 \\
\texttt{Claude-4-Sonnet} & - & 58.2 $\pm$ 2.1& 62.9 $\pm$ 1.7 & 61.7 $\pm$ 4.3& \underline{77.5 $\pm$ 1.4}& 32.8 $\pm$ 1.9& 58.6 $\pm$ 2.3 \\
\texttt{GPT-4.1} & - & 52.2 $\pm$ 1.5& 63.4 $\pm$ 2.6 & 67.4 $\pm$ 2.3& 70.0 $\pm$ 1.9& 55.5 $\pm$ 3.1& 61.7 $\pm$ 2.2 \\
\texttt{Gemini-2.5-Flash} & - & 25.1 $\pm$ 1.3& 67.3 $\pm$ 1.6 & 28.1 $\pm$ 2.1& 24.1 $\pm$ 1.8& 24.9 $\pm$ 1.8& 33.9 $\pm$ 1.7 \\

\midrule
\multicolumn{8}{c}{\textbf{Tool-Instruction Methods}} \\
\midrule
\texttt{Graph-ToolFormer} & 8B & 47.5 $\pm$ 1.9& 68.1 $\pm$ 4.8& 74.7 $\pm$ 4.2& 61.4 $\pm$ 3.4& 65.8 $\pm$ 4.5& 62.4 $\pm$ 3.5\\
\texttt{GraphForge} & 8B & 63.5 $\pm$ 3.5& \underline{70.9 $\pm$ 5.4}& \underline{80.4 $\pm$ 4.2}& 63.4 $\pm$ 4.4& \underline{73.5 $\pm$ 3.1}& \underline{70.2 $\pm$ 3.8}\\

\texttt{ToolGen} & 8B & \underline{75.8 $\pm$ 1.1}& 57.9 $\pm$ 2.9& 79.4 $\pm$ 2.3& 61.2 $\pm$ 1.3& 62.7 $\pm$ 1.5& 67.4 $\pm$ 1.8\\
\midrule
\texttt{GraphChain} & 7B & \textbf{81.5 $\pm$ 2.2} & \textbf{81.1 $\pm$ 0.7} & \textbf{89.6 $\pm$ 2.0} & \textbf{83.6 $\pm$ 2.6} & \textbf{84.1 $\pm$ 0.3} & \textbf{84.7 $\pm$ 1.8} \\
\midrule
\textit{Relative improvement (\%)} & - & \textit{+7.5\%} & \textit{+14.4\%} & \textit{+11.4\%} & \textit{+7.9\%} & \textit{+14.4\%} & \textit{+20.7\%} \\
\bottomrule
\multicolumn{8}{r}{\scriptsize{$^\dagger$ \textbf{Boldface} denotes the highest score, and \underline{underline} indicates the best result among baselines.}}
\end{tabular}%
}
\end{table}

Table~\ref{tab:main_results} presents performance comparisons of \texttt{GraphChain} against state-of-the-art baselines, with statistical significance confirmed by two-sample t-tests ($p < 0.05$). Key insights include:

\begin{itemize}[leftmargin=*]
    \item \texttt{GraphChain} substantially outperforms all baselines, achieving 84.7\% average accuracy compared to 70.2\% for the best baseline (\texttt{GraphForge}), representing a 20.7\% relative improvement.
    
    \item Among text-instruction baselines, \texttt{GPT-4o} with approximately 200B parameters demonstrates superior performance (59.4\% average accuracy), confirming the applicability of scaling laws to graph reasoning tasks.
    
    \item Specialized graph reasoning approaches like \texttt{GraphForge} (70.2\% average accuracy) significantly outperform even the largest general-purpose LLMs.
    
    \item \texttt{GraphChain} achieves these results with only 7B parameters, compared to \texttt{GraphForge}'s 8B and \texttt{GPT-4o}'s ~200B, demonstrating remarkable parameter efficiency.
\end{itemize}

\subsection{Ablation Study}
\begin{figure}[h]
\centering
\begin{minipage}{0.44\textwidth}
   \centering
   \includegraphics[width=\textwidth]{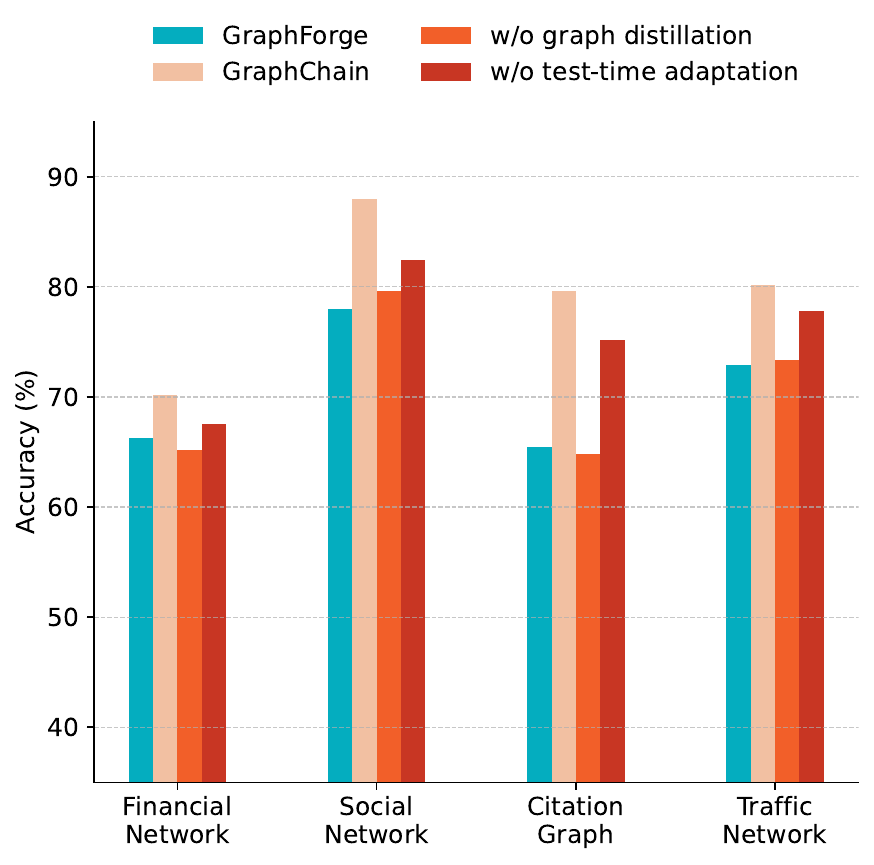}
   \caption{\small Impact of removing graph distillation or test-time adaptation.}
   \label{fig:ablation}
\end{minipage}
\hfill
\hspace{-2ex}
\begin{minipage}{0.54\textwidth}
   \centering
   \includegraphics[width=\textwidth]{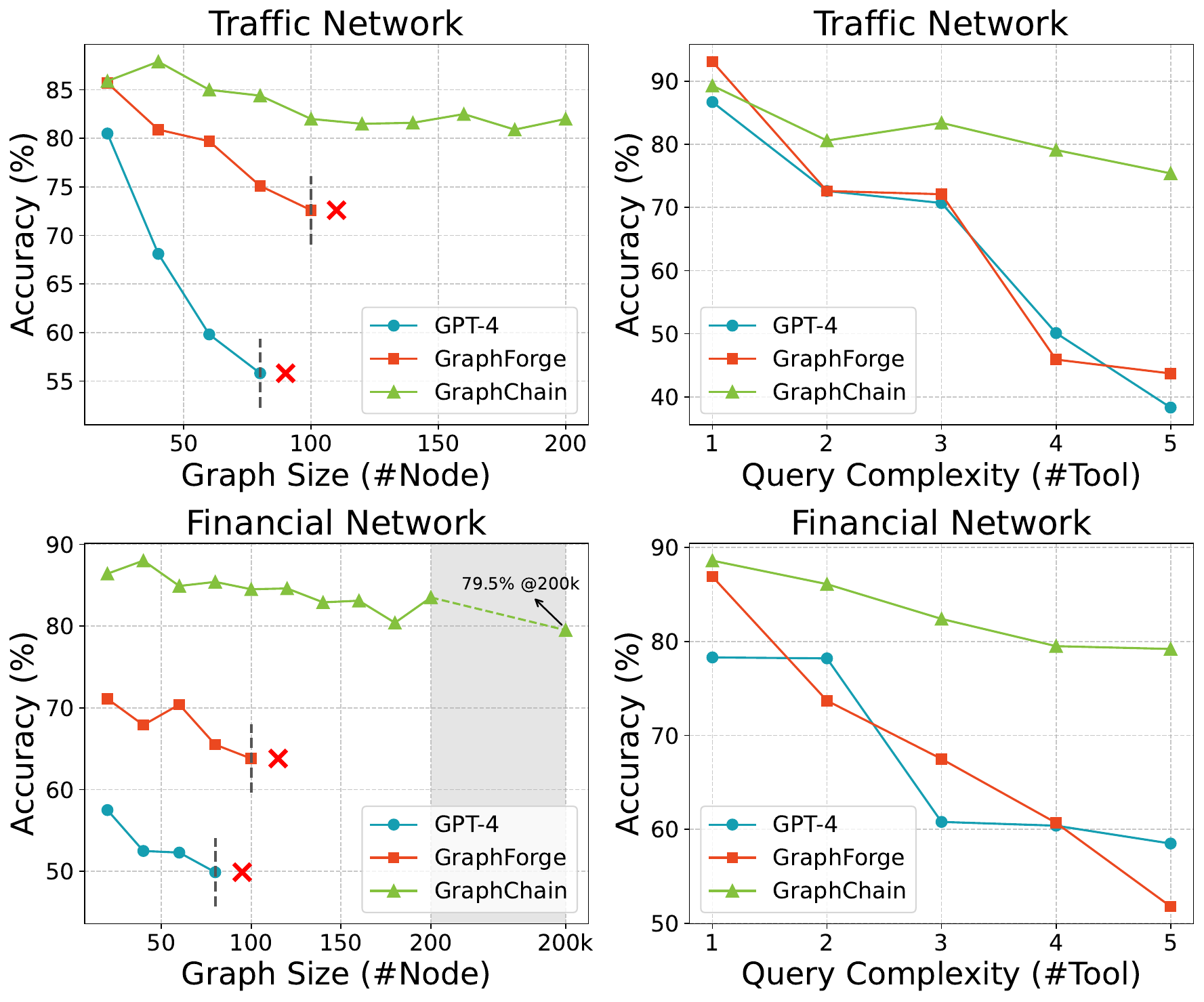}
   \caption{\small Comparison with varying Graph Sizes and Query Complexity.}
   \label{fig:complexity}
\end{minipage}
\end{figure}

We introduced two variants: (1) \textbf{w/o graph distillation}, where the progressive graph distillation mechanism is disabled; and (2) \textbf{w/o test-time adaptation}, where the Structure-aware Test-Time Adaptation (STTA) component is removed during inference.
Figure~\ref{fig:ablation} reveals several key insights:

\underline{First}, \texttt{GraphChain} consistently outperforms \texttt{GraphForge} across all graph scenarios, demonstrating the superiority of our approach. \underline{Second}, Removing either component leads to performance degradation, confirming that both play critical roles in enabling effective tool-chaining and structural understanding. \underline{Third}, The performance drop is more severe when graph distillation is removed compared to when disabling STTA, highlighting that progressive distillation is particularly crucial for graph analysis. \underline{Lastly}, \texttt{GraphChain} without test-time adaptation still outperforms \texttt{GraphForge} in most scenarios, indicating that our multi-step tool-chaining approach with graph distillation is inherently more effective than single-step tool invocation patterns.

\subsection{Scalability Analysis}

\label{subsec:scalibility}
We stratified our evaluation dataset based on graph size (node count) and reasoning complexity (tool sequence length) to assess how performance scales with these dimensions. Figure~\ref{fig:complexity} reveals:

(1). As graph size increases, baselines exhibit significant performance degradation, with \texttt{GPT-4o} declining more dramatically, demonstrating the limitations of text-instruction for larger graphs.
   
(2). \texttt{GraphChain} maintains its performance advantage consistently across all graph sizes tested, including graphs with up to 200,000 nodes. This exceptional scalability stems from representing memory states through concise natural language summaries rather than direct graph descriptions.
   
(3). While all methods perform well on simple queries (requiring 1-2 tool calls), performance disparities increase with query complexity. Both \texttt{GPT-4o} and \texttt{GraphForge} show steep declines for queries requiring 4-5 tool calls, while \texttt{GraphChain} maintains higher accuracy, demonstrating superior capability for multi-step reasoning.

\subsection{Transfer Learning Evaluation}

To assess transfer capabilities, we fine-tuned \texttt{GraphChain} exclusively on Financial Network and evaluated on three unseen domains, comparing performance with and without the STTA module.

\begin{table}[h]
\centering
\caption{\small Results (accuracy \%) when training on Financial Network and testing on other domains.}
\label{tab:transfer}
\resizebox{0.6\textwidth}{!}{%
\begin{tabular}{lccc}
\toprule
\textbf{Model} & \textbf{Social Network} & \textbf{Citation Graph} & \textbf{Traffic Network} \\
\midrule
\texttt{GraphChain} (in-domain) & 89.6 & 83.6 & 84.1 \\
\texttt{GraphChain} w/ STTA & 86.8 (-3.1\%) & 79.2 (-4.3\%) & 80.3 (-4.5\%) \\
\texttt{GraphChain} w/o STTA & 84.5 (-5.7\%) & 75.1 (-10.2\%) & 77.4 (-8.0\%) \\
\bottomrule
\end{tabular}
}
\end{table}

Results in Table~\ref{tab:transfer} demonstrate \texttt{GraphChain}'s strong transfer learning capabilities, with cross-domain performance closely approaching in-domain results. The STTA mechanism substantially improves transfer performance, reducing accuracy drops by 2.6\%, 5.9\%, and 3.5\% across the three target domains compared to the variant without STTA, confirming its effectiveness in adapting to diverse graph structures without domain-specific retraining.

\subsection{Tool Chain Analysis}

To understand how \texttt{GraphChain} adapts its exploration strategy across domains, we categorized tools into six functional clusters and analyzed their usage patterns.

\begin{figure}[h]
\centering
\includegraphics[width=0.8\textwidth]{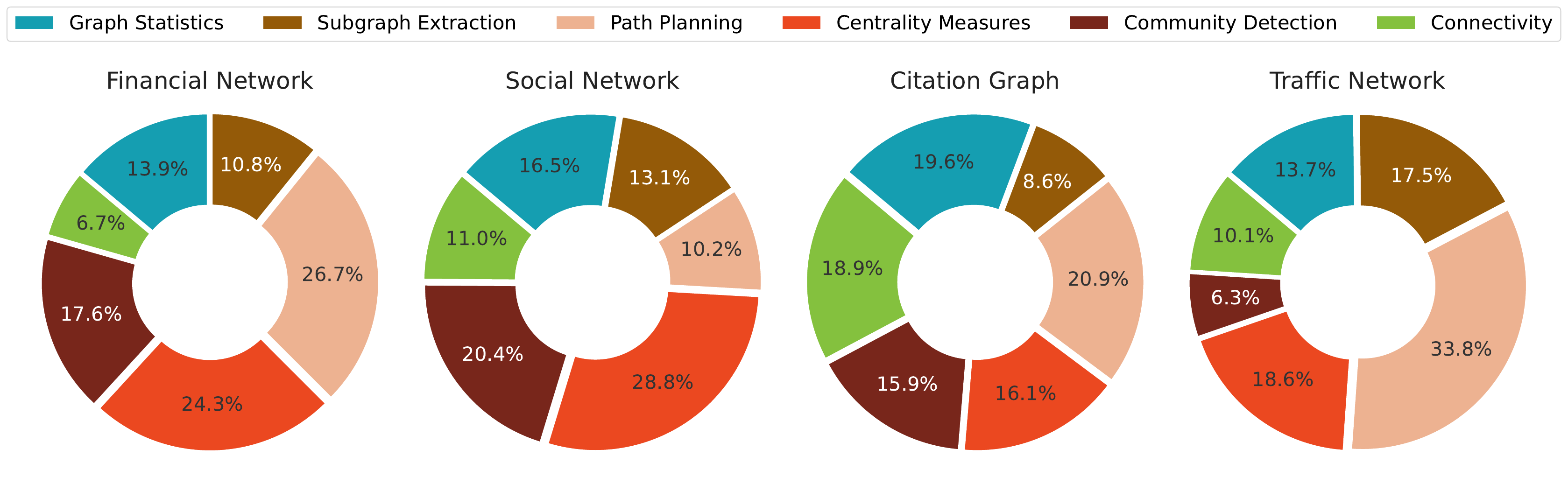}
\caption{\small Distribution of tool types utilized by \texttt{GraphChain} across different graph domains.}
\label{fig:tool_distribution}
\end{figure}

Figure~\ref{fig:tool_distribution} reveals distinct exploration patterns adapted to each domain's characteristics. Path Planning tools dominate in Traffic Network (33.8\%) and Financial Network (26.7\%), reflecting the importance of traversal analysis. Social Network analysis relies on Centrality Measures (28.8\%) and Community Detection (20.4\%), aligning with the importance of influence and clustering. Citation Graph processing shows a more balanced distribution with significant usage of Connectivity tools (18.9\%). 
These domain-specific variations demonstrate \texttt{GraphChain}'s ability to adaptively construct appropriate tool sequences on different graph scenarios.

Our framework is designed to be inherently robust to variations in the tool library. The core of \texttt{GraphChain} employs a reinforcement learning policy that learns to select optimal tool sequences from the available action space, rather than being hard-coded to specific tools.To empirically validate this robustness, we conducted an additional experiment with a reduced toolset (removing 50\% of tools from Centrality and Community Detection categories). Table~\ref{tab:robustness} demonstrate that \texttt{GraphChain} maintains strong performance even with a reduced toolset, showcasing the adaptability of our RL agent in finding alternative tool sequences to solve tasks.

\begin{table}[h]
\centering
\caption{\small Robustness to Tool Library Composition (accuracy \%).}
\label{tab:robustness}
\resizebox{0.8\textwidth}{!}{%
\begin{tabular}{lcccccc}
\toprule
\thead{\textbf{Model}} & \thead{\textbf{Financial}\\ \textbf{Network}} & \thead{\textbf{Chemical}\\ \textbf{Molecule}} & \thead{\textbf{Social}\\ \textbf{Network}} & \thead{\textbf{Citation}\\ \textbf{Graph}} & \thead{\textbf{Traffic}\\ \textbf{Network}} & \textbf{Average} \\
\midrule
\texttt{GraphForge} (Baseline) & 63.5$\pm$3.5 & 70.9$\pm$5.4 & 80.4$\pm$4.2 & 63.4$\pm$4.4 & 73.5$\pm$3.1 & 70.2$\pm$3.8 \\
\texttt{GraphChain} (Full Toolset) & \textbf{81.5$\pm$2.2} & \textbf{81.1$\pm$0.7} & \textbf{89.6$\pm$2.0} & \textbf{83.6$\pm$2.6} & \textbf{84.1$\pm$0.3} & \textbf{84.7$\pm$1.8} \\
\texttt{GraphChain} (Reduced Toolset) & 77.3$\pm$2.8 & 78.4$\pm$1.8 & 82.7$\pm$3.1 & 80.1$\pm$3.2 & 80.6$\pm$0.9 & 79.8$\pm$2.4 \\
\bottomrule
\end{tabular}
}
\end{table}

\subsection{Robustness Study}

In order to validate the robustness of \texttt{GraphChain}, we evaluated \texttt{GraphChain} with different base models. The results are shown in Table~\ref{tab:rob_base_model}.



\begin{table}[h]
  \centering
  \caption{\small Performance with Different Base LLMs (accuracy \%)}
  \label{tab:rob_base_model}
  \resizebox{0.7\textwidth}{!}{
  \begin{tabular}{lcccccc}
    \toprule
    \thead{\textbf{Base Model}} & \thead{\textbf{Financial}\\ \textbf{Network}} & \thead{\textbf{Chemical}\\ \textbf{Molecule}} & \thead{\textbf{Social}\\ \textbf{Network}} & \thead{\textbf{Citation}\\ \textbf{Graph}} & \thead{\textbf{Traffic}\\ \textbf{Network}} & \textbf{Average} \\
    \midrule
    Qwen2.5-7B & 70.5 & 81.1 & 90.4 & 79.0 & 82.0 & 80.6 \\
    Llama3.1-8B & 69.3 & 81.7 & 93.7 & 82.5 & 81.7 & 81.8 \\
    GLM4-9B & 70.2 & 78.9 & 93.8 & 79.7 & 79.9 & 80.5 \\
    \bottomrule
  \end{tabular}
  }
\end{table}

The consistent superior results across different base models demonstrate the robustness and general applicability of our approach. We also conducted supplementary experiments using Qwen2.5 models of varying sizes (3B, 7B, and 14B). The Table~\ref{tab:rob_size} show that \texttt{GraphChain}'s performance improves with larger model sizes. Notably, even the smaller 3B model still maintain reasonable performance under our framework.


\begin{table}[h]
  \centering
  \caption{Comparison of Base Models with Different Sizes (accuracy \%)}
  \label{tab:rob_size}
  \resizebox{0.6\textwidth}{!}{
  \begin{tabular}{lccccc}
    \toprule
    \thead{\textbf{Model}\\ \textbf{Size}} & \thead{\textbf{Financial}\\ \textbf{Network}} & \thead{\textbf{Chemical}\\ \textbf{Molecule}} & \thead{\textbf{Social}\\ \textbf{Network}} & \thead{\textbf{Citation}\\ \textbf{Graph}} & \thead{\textbf{Traffic}\\ \textbf{Network}} \\
    \midrule
    Qwen2.5-3B & 63.1\% & 56.9\% & 70.2\% & 74.4\% & 73.4\% \\
    Qwen2.5-7B & 81.5\% & 81.1\% & 89.6\% & 83.6\% & 84.1\% \\
    Qwen2.5-14B & 85.7\% & 85.4\% & 92.2\% & 83.2\% & 89.7\% \\
    \bottomrule
  \end{tabular}
  }
\end{table}

%% file: data/conclusion.tex
\section{Conclusions and Limitation}
\label{sec:conclusion}

In this paper, we introduced GraphChain, a novel framework that enables LLMs to effectively process and reason over large-scale graph data through dynamic tool-chaining. By integrating progressive graph distillation with structure-aware test-time adaptation, GraphChain addresses the fundamental challenges of context exhaustion and reasoning hallucination that plague existing graph processing approaches. Our extensive experiments across diverse domains demonstrate that GraphChain significantly outperforms prior methods. 

Our current implementation primarily focuses on static graphs and may require adaptation for dynamic or temporal graph structures that evolve over time. The tool library used in our experiments, though comprehensive, could be expanded to include more domain-specific operations for specialized applications. These limitations present valuable directions for future research.



%% file: app/proof_of_IB.tex
\section{Proof of Proposition~\ref{prop:ib_compression}}
\label{app:proof_ib_prop}

We start with the fundamental assumptions:
\begin{enumerate}
    \item The input $X$ is generated from underlying factors including task-relevant information $Y$ and task-irrelevant information $IR$.
    \item The process forms a Markov chain: $(Y, IR) \rightarrow X \rightarrow \mathbf{m}_t$. This signifies that the memory state $\mathbf{m}_t$ is generated based on the input $X$, which itself is derived from the underlying factors $(Y, IR)$.
    \item The optimization objective derived from the reward function $R_t$ (Eq.~\ref{eq:unified_reward_distill}) encourages policies that produce trajectories where intermediate states $\mathbf{m}_t$ have high task relevance $\mathrm{Rel}(\mathbf{m}_t, \mathcal{Q})$ and low complexity/volume $\mathrm{GDL}(\mathbf{m}_t)$.
    \item Based on the proposition's statement, maximizing relevance correlates with maximizing $I(Y; \mathbf{m}_t)$, and minimizing GDL correlates with minimizing the overall information captured from the input, $I(X; \mathbf{m}_t)$.
\end{enumerate}

According to the Data Processing Inequality (DPI)~\citep{beaudry2011intuitive} applied to the Markov chain $(Y, IR) \rightarrow X \rightarrow \mathbf{m}_t$, the information that the final representation $\mathbf{m}_t$ retains about the initial factors $(Y, IR)$ cannot exceed the information it retains about the intermediate variable $X$:
\begin{equation}
    I((Y, IR); \mathbf{m}_t) \le I(X; \mathbf{m}_t)
\label{eq:proof_dpi}
\end{equation}

Now, we apply the chain rule for mutual information to the term on the left-hand side:
\begin{equation}
    I((Y, IR); \mathbf{m}_t) = I(Y; \mathbf{m}_t) + I(IR; \mathbf{m}_t | Y)
\label{eq:proof_chain_rule_revised}
\end{equation}
Here, $I(Y; \mathbf{m}_t)$ represents the information that the memory state $\mathbf{m}_t$ contains about the relevant variable $Y$. The term $I(IR; \mathbf{m}_t | Y)$ represents the \textit{additional} information that $\mathbf{m}_t$ contains about the irrelevant variable $IR$, given that the relevant information $Y$ is already known. This term quantifies the amount of irrelevant information captured by $\mathbf{m}_t$ beyond what is already explained by its correlation with $Y$.

Substituting the expansion from Eq.~\ref{eq:proof_chain_rule_revised} into the DPI (Eq.~\ref{eq:proof_dpi}), we obtain:
\begin{equation}
    I(Y; \mathbf{m}_t) + I(IR; \mathbf{m}_t | Y) \le I(X; \mathbf{m}_t)
\label{eq:proof_combined_inequality}
\end{equation}

Rearranging this inequality gives us an upper bound on the conditional mutual information involving the irrelevant component:
\begin{equation}
    I(IR; \mathbf{m}_t | Y) \le I(X; \mathbf{m}_t) - I(Y; \mathbf{m}_t)
\label{eq:proof_irrelevant_info_bound}
\end{equation}
This inequality shows that the amount of irrelevant information retained in $\mathbf{m}_t$ (conditioned on the relevant part $Y$) is upper-bounded by the difference between the total information $\mathbf{m}_t$ captures from the input $X$ and the useful information it captures about the target $Y$.

Now, let's consider the optimization objective implied by the progressive distillation reward function (Eq.~\ref{eq:unified_reward_distill}). This objective aims to find a policy $\pi_\phi$ that maximizes the expected return. The reward structure encourages steps that increase relevance (proxy for $I(Y; \mathbf{m}_t)$) and decrease GDL (proxy for $I(X; \mathbf{m}_t)$). Thus, the optimization process implicitly seeks intermediate states $\mathbf{m}_t$ that achieve a trade-off, aiming for:
\begin{itemize}
    \item High $I(Y; \mathbf{m}_t)$ (maximize information about the answer)
    \item Low $I(X; \mathbf{m}_t)$ (minimize total information stored, i.e., compress the input)
\end{itemize}
By striving to maximize $I(Y; \mathbf{m}_t)$ and minimize $I(X; \mathbf{m}_t)$, the optimization process is effectively attempting to minimize the upper bound $I(X; \mathbf{m}_t) - I(Y; \mathbf{m}_t)$ found in Eq.~\ref{eq:proof_irrelevant_info_bound}.

Minimizing this upper bound serves as a direct mechanism to suppress the amount of irrelevant information $I(IR; \mathbf{m}_t | Y)$ contained in the memory state $\mathbf{m}_t$. Therefore, optimizing the distillation-aware reward function encourages the policy to generate representations that form an information bottleneck, preserving task-relevant information while actively discarding task-irrelevant information, as quantified by the bound derived from the Data Processing Inequality.

%% file: app/case_study.tex
\section{Case Study}

\begin{figure}[h]
    \centering
\includegraphics[width=0.99\textwidth]{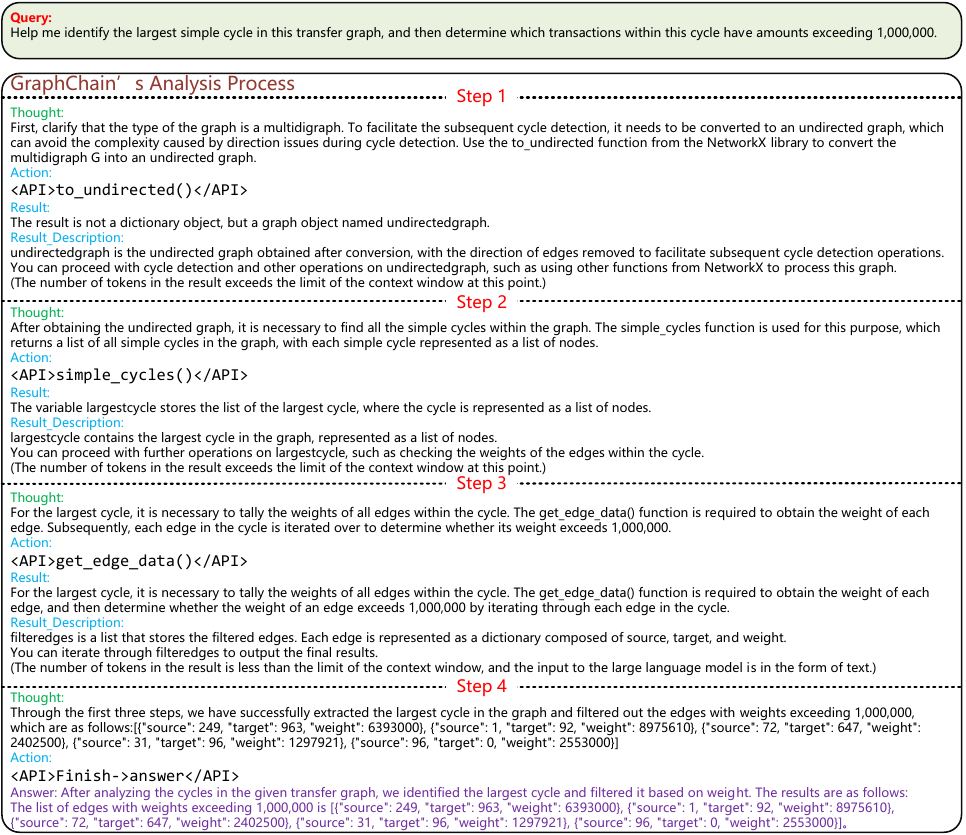}
    \caption{\small A typical case of \texttt{GraphChain} on Financial Networks.}
    \label{fig:case_study}
\end{figure}

We present a case study in Figure~\ref{fig:case_study} involving a transmission network from the Elliptic Dataset, comprising 11,896 nodes and 31,054 edges—exceeding the context processing capabilities of current LLMs. 
\texttt{GraphChain} demonstrates robust capability to systematically handle this task through a series of tool invocations. The Progressive Graph Distillation mechanism enables iterative reduction of information complexity, focusing on the most salient features.

Due to context window limitations, we store intermediate processing results in memory rather than directly inputting them into the LLM. The model outputs detailed descriptions of current processing results and their usage methods to facilitate further analysis. Only when intermediate results' token count falls below the maximum context window limit are they directly input into the LLM, maximizing the model's advantages while minimizing limitations.

%% file: app/experiment_setup.tex
\section{Details of Experimental Setup}
\label{app:details_of_exp}

We provide comprehensive details on our experimental setup to ensure reproducibility. All experiments were conducted on 2 NVIDIA A800 80GB GPUs, using LoRA-based fine-tuning (rank $r$=16, alpha=32) on the \texttt{Qwen2.5-7B-instruction} model.

\subsection{Training Configuration}

Our training pipeline consisted of three main stages:

\begin{itemize}[leftmargin=*]
   \item \textbf{Supervised Fine-Tuning (SFT) Stage:} We used a learning rate of $5 \times 10^{-5}$ with 4\% warmup and a cosine scheduler for 8 epochs. This initial phase established the model's ability to follow graph reasoning instructions.
   
   \item \textbf{Reinforcement Learning (RL) Stage:} We implemented Proximal Policy Optimization (PPO) with step-level rewards, departing from traditional RLHF approaches that apply rewards solely to the final step. Our implementation used:
   \begin{itemize}
       \item Learning rate: $1 \times 10^{-5}$
       \item Batch size: 8
       \item Initial KL coefficient: 0.3
       \item Loss coefficient ($\beta$): 0.15
       \item GAE parameter ($\lambda$): 0.95
       \item Discount factor ($\gamma$): 0.99
   \end{itemize}
   
   \item \textbf{Test-Time Adaptation Stage:} For the structure-aware adaptation mechanism, we configured:
   \begin{itemize}
       \item Learning rate: 0.01
       \item Batch size: 10
   \end{itemize}
\end{itemize}

\subsection{Inference Settings}

During inference, we used a temperature of 0.7 and top-p value of 1.0, optimizing for a balance between diversity and coherence in the generated tool chains.

Complete configuration files and scripts are available in our code repository to facilitate reproduction of our results.

%% file: app/baseline_app.tex
\section{Baseline Implementation}
\label{app:baseline}

To comprehensively evaluate \texttt{GraphChain}, we implemented several state-of-the-art graph reasoning baselines spanning both Text-Instruction and Tool-Instruction paradigms. Table~\ref{tab:baselines} summarizes these baseline methods and their corresponding backbone models.

\begin{table}[t]
\centering
\caption{\small Comparison of baseline methods and their corresponding models for graph reasoning.}
\vspace{0.5em}
\renewcommand{\arraystretch}{1.3}
\resizebox{\textwidth}{!}{%
\begin{tabular}{lcll}
\toprule
\textbf{LLM Type} & \textbf{Open Source} & \textbf{Method} & \textbf{Base Model} \\
\midrule
\multirow{4}{*}{Text Instruction} & \ding{55} & Two-shot & \texttt{Claude-series}~\citep{TheC3} \\
& \ding{55} & Two-shot & \texttt{GPT-series}~\citep{DBLP:journals/corr/abs-2303-08774} \\
\cmidrule(lr){2-4}
& \checkmark & \texttt{NLGraph}~\citep{DBLP:conf/nips/WangFHTHT23} & \texttt{GPT-4-turbo} \\
& \checkmark & \texttt{GraphWiz}~\citep{DBLP:journals/corr/abs-2402-16029} & \texttt{Llama2-13B} \\
\midrule
\multirow{4}{*}{Tool Instruction} & \ding{55} & Function Calling & \texttt{GPT-3.5-turbo}~\citep{DBLP:journals/corr/abs-2303-08774} \\
& \ding{55} & Function Calling & \texttt{GPT-4o}~\citep{DBLP:journals/corr/abs-2303-08774} \\
& \ding{55} & Function Calling & \texttt{GLM4-0520}~\citep{glm2024chatglm} \\
\cmidrule(lr){2-4}
& \checkmark & \texttt{Graph-ToolFormer}~\citep{DBLP:journals/corr/abs-2304-11116} & \texttt{Llama3-8B} \\
& \checkmark & \texttt{GraphForge}~\citep{wang2024graphtool} & \texttt{Llama3-8B} \\
\bottomrule
\end{tabular}%
}
\label{tab:baselines}
\end{table}

\subsection{Text-Instruction Methods}

For closed-source LLMs (\texttt{Claude-series}, \texttt{GPT-series}, and \texttt{GLM4-0520}), we employed two-shot Chain-of-Thought (CoT) prompting to stimulate structured reasoning. Table~\ref{tab:two-shot-example} illustrates our prompting approach with an example.

\begin{table}[t]
\centering
\caption{\small Example of two-shot prompting used for text-instruction baselines.}
\label{tab:two-shot-example}
\begin{tabular}{p{0.95\textwidth}}
\toprule
\textbf{Prompt Template} \\
\midrule
\small
\textit{You are an AI assistant specialized in graph reasoning. For each problem, first extract the graph structure, then solve the task step by step.}

\textbf{\textcolor{blue}{Example 1:}}\\
\textbf{Input:} Given a weighted directed graph with edges: \texttt{[(0, 2, \{'weight': 3\}), (0, 3, \{'weight': 7\}), (1, 0, \{'weight': 2\}), (1, 4, \{'weight': 8\}), (2, 4, \{'weight': 1\}), (3, 4, \{'weight': 3\})]}. Find the shortest path from node 1 to node 4.

\textbf{Reasoning:} First, I'll extract the graph structure.\\
\texttt{Graph: [(0, 2, \{'weight': 3\}), (0, 3, \{'weight': 7\}), (1, 0, \{'weight': 2\}), (1, 4, \{'weight': 8\}), (2, 4, \{'weight': 1\}), (3, 4, \{'weight': 3\})]}

Now I'll find the shortest path from node 1 to node 4.
\begin{itemize}
\item Direct path: \texttt{1 → 4} with weight \texttt{8}
\item Indirect path: \texttt{1 → 0 → 2 → 4} with weight \texttt{2 + 3 + 1 = 6}
\item Indirect path: \texttt{1 → 0 → 3 → 4} with weight \texttt{2 + 7 + 3 = 12}
\end{itemize}
The shortest path is \texttt{1 → 0 → 2 → 4} with total weight \texttt{6}.

\textbf{Answer:} \textcolor{purple}{The shortest path from node 1 to node 4 is \texttt{1 → 0 → 2 → 4} with total weight 6.}\\

\textbf{\textcolor{blue}{Example 2:}}\\
\textbf{Input:} Given an undirected graph with edges: \texttt{[(0, 1), (0, 3), (1, 2), (1, 4), (2, 5), (3, 4), (4, 5)]}. Determine if there exists a cycle in this graph.

\textbf{Reasoning:} Let me extract the graph structure.\\
\texttt{Graph: [(0, 1), (0, 3), (1, 2), (1, 4), (2, 5), (3, 4), (4, 5)]}

To check for cycles, I'll trace possible paths:\\
Starting from node 0: \texttt{0 → 1 → 2 → 5 → 4 → 3 → 0}\\
This forms a cycle: \texttt{0 → 1 → 2 → 5 → 4 → 3 → 0}

\textbf{Answer:} \textcolor{purple}{Yes, there exists a cycle in the graph.}\\

\textbf{\textcolor{red}{Now solve the following problem:}}\\
\textbf{Input:} [Problem description]
\\
\bottomrule
\end{tabular}
\end{table}

For specialized graph reasoning methods, we reproduced:

\begin{itemize}[leftmargin=*]
   \item \texttt{GraphWiz}: We reimplemented using the \texttt{Llama2-13B} backbone with a learning rate of 2e-5 for 3 epochs and maximum sequence length of 2048. During inference, temperature was set to 0 with maximum output tokens of 1024 to ensure stable generation.
   
   \item \texttt{NLGraph}: Following the original implementation, we provided 4 exemplars for connectivity and cycle tasks, and 5 exemplars for other tasks due to context size limitations. For fair comparison, we used the standardized test set across all experiments.
\end{itemize}

\subsection{Tool-Instruction Methods}

We implemented tool-augmented approaches including:

\begin{itemize}[leftmargin=*]
   \item \texttt{Graph-ToolFormer}: We reimplemented this approach based on the \texttt{Llama3-8B} model using LoRA (rank $r$=16, alpha=32) with a learning rate of 1e-5 and weight decay of 1e-2 for 3 epochs. For generation, we used beam search with 5 beams, top-k of 5, top-p of 0.95, and temperature of 0.7.
   
   \item \texttt{GraphForge}: We implemented based on \texttt{Llama3-8B} using LoRA (rank $r$=16, alpha=32) with a learning rate of 5e-5 for 5 epochs. Inference settings matched our \texttt{GraphChain} configuration with temperature of 0.7 and top-p of 1.0.
   
   \item Function Calling: For closed-source models supporting function calling (\texttt{GPT-3.5-turbo}, \texttt{GPT-4o}, and \texttt{GLM4-0520}), we implemented the same graph processing functions used in \texttt{GraphChain} as external API tools, allowing these models to leverage structured tool invocation capabilities during inference.
\end{itemize}

All baseline implementations were executed using the same hardware setup as \texttt{GraphChain}: two NVIDIA A800 GPUs for fine-tuning and inference with open-source models. For closed-source models, we utilized their respective official API interfaces. To ensure fair comparison across all methods, we partitioned original graphs into subgraphs with fewer than 100 nodes for evaluation, while separately testing \texttt{GraphChain}'s scalability on full-sized graphs with up to 200,000 nodes in Section~\ref{subsec:scalibility}.

%% file: app/nx_function.tex
\section{Graph Analysis Tool Library}
\label{app:tool_library}

\begin{table}[t!]
\centering
\caption{NetworkX Functions Categorized by Graph Analysis Task}
\label{tab:nx_functions}
\begin{tabular}{|>{\columncolor{gray!10}}p{4cm}|>{\raggedright\arraybackslash}p{8cm}|} 
\toprule
\rowcolor{gray!25}
\textbf{Category} & \textbf{NetworkX Functions} \\ 
\midrule
Basic Graph Properties & 
\texttt{G.number\_of\_nodes()}, 
\texttt{G.number\_of\_edges()},
\texttt{G.has\_node(n)},
\texttt{G.has\_edge(u, v)},
\texttt{G.degree()},
\texttt{G.in\_degree()},
\texttt{G.out\_degree()},
\texttt{G.get\_edge\_data(u, v)} \\ 
\midrule
Centrality Metrics & 
\texttt{nx.betweenness\_centrality()},
\texttt{nx.closeness\_centrality()},
\texttt{nx.degree\_centrality()},
\texttt{nx.eigenvector\_centrality()},
\texttt{nx.harmonic\_centrality()},
\texttt{nx.percolation\_centrality()},
\texttt{nx.second\_order\_centrality()},
\texttt{nx.subgraph\_centrality()} \\ 
\midrule
Connectivity and Components & 
\texttt{nx.strongly\_connected\_components()},
\texttt{nx.weakly\_connected\_components()},
\texttt{nx.articulation\_points()},
\texttt{nx.bridges()},
\texttt{nx.k\_edge\_components()},
\texttt{nx.k\_node\_components()},
\texttt{nx.node\_connectivity()},
\texttt{nx.edge\_connectivity()} \\ 
\midrule
Shortest Paths and Distances & 
\texttt{nx.all\_pairs\_shortest\_path()},
\texttt{nx.all\_pairs\_shortest\_path\_length()},
\texttt{nx.dijkstra\_path()},
\texttt{nx.dijkstra\_path\_length()},
\texttt{nx.floyd\_warshall()} \\ 
\midrule
Clustering and Communities & 
\texttt{nx.average\_clustering()},
\texttt{nx.clustering()},
\texttt{nx.transitivity()},
\texttt{nx.triangles()},
\texttt{nx.label\_propagation\_communities()},
\texttt{nx.louvain\_communities()} \\ 
\midrule
Flow Algorithms & 
\texttt{nx.boykov\_kolmogorov\_min\_cut()},
\texttt{nx.dinic\_min\_cut()},
\texttt{nx.edmonds\_karp\_min\_cut()},
\texttt{nx.minimum\_cut()} \\ 
\midrule
Cycle Detection & 
\texttt{nx.simple\_cycles()},
\texttt{nx.cycle\_basis()} \\ 
\midrule
Topological Sorting & 
\texttt{nx.topological\_sort()},
\texttt{nx.is\_directed\_acyclic\_graph()},
\texttt{nx.all\_topological\_sorts()},
\texttt{nx.topological\_generations()} \\ 
\bottomrule
\end{tabular}
\end{table}

To construct an effective graph question-answering system, we selected 45 functions from the NetworkX library through a systematic review of graph analysis tasks prevalent in academic research and practical applications. Table~\ref{tab:nx_functions} shows the complete list of selected functions. The selection process prioritized coverage of eight core dimensions of graph analytics:

\begin{itemize}[leftmargin=*]
   \item \texttt{Basic Graph Properties} -- Functions providing structural metadata, including node/edge counts, degree distributions, and adjacency queries. 
   \item \texttt{Centrality Metrics} -- Measures for node influence, spanning degree centrality to advanced methods (eigenvector, percolation, and Katz centrality).
   \item \texttt{Connectivity and Components} -- Tools for evaluating graph robustness, such as articulation points, bridges, and strongly/weakly connected components.
   \item \texttt{Shortest Paths and Distances} -- Algorithms for unweighted and weighted paths, critical for routing and diffusion modeling.
   \item \texttt{Clustering and Communities} -- Modular structure analysis via clustering coefficients and detection algorithms (e.g., label propagation, Louvain).
   \item \texttt{Flow Algorithms} -- Maximum flow and minimum cut computations using multiple methodologies (e.g., Edmonds-Karp).
   \item \texttt{Cycle Analysis} -- Feedback loop identification in directed and undirected graphs.
   \item \texttt{Topological Sorting} -- Dependency resolution for directed acyclic graphs (DAGs).
\end{itemize}

While not exhaustive, this set was carefully selected to balance \textit{analytical breadth} and \textit{computational efficiency}, ensuring system responsiveness and interpretability. Future work may integrate domain-specific or higher-order analytics, but this toolset is representative and sufficient for general-purpose graph analysis.

%% file: app/dataset.tex
\section{Data Construction}
\label{app:data_construction}

\begin{table}[t!]
\centering
\caption{\small The prompt template for constructing the SFT dataset.}
\label{tab:prompt_for_SFT}
\begin{tabular}{|>{\columncolor{gray!10}}p{2cm}|p{10cm}|}
\toprule
\rowcolor{gray!25}
\textbf{Category} & \textbf{Description} \\ 
\midrule
Dataset Name & \texttt{Citation-Network.txt} \\ 
\midrule
Dataset Type & \texttt{MultiDirected Graph} \\ 
\midrule
Dataset Content & \small The citation data between research papers. Directed edge A to B means that paper A cites paper B. The graph construction operation is: \\
& \texttt{\textcolor{blue}{G = nx.MultiDiGraph(), G.add\_edge(paper1, paper2)}}, \\
& where \texttt{paper1} and \texttt{paper2} are research papers. String type is used to store nodes. \\ 
\midrule
Task & \small Generate a complex graph problem and its step-by-step solution process. \\ 
\midrule
Output Type & \small \texttt{JSON} \\ 
\midrule
Output Rules & \small \textbf{(1)} The output must be a JSON containing a series of "from" and "value" as shown in the example, using English. \\
& \textbf{(2)} Provide the user problem in "value" under "user", generate the response in "value" under "assistant", and generate API return results in "value" under "function". \\
& \textbf{(3)} The output can have only this JSON data with no additional information. \\
& \textbf{(4)} Follow the format of the example but exclude the key name "example". \\ 
\midrule
\rowcolor{gray!10}
Special Attention & \small In the "assistant" response, provide detailed thought processes without code, using NetworkX methods. Mark called APIs with \texttt{\textcolor{green}{<API>...</API>}}, e.g., \texttt{\textcolor{green}{<API>nx.dfs\_edges(graph, source=10)</API>}}. Format outputs as: \\
& \texttt{\textcolor{blue}{Thought: ... Action: <API>...</API>}} \\
& When \texttt{<API>...</API>} is encountered, provide "function" with fabricated API results: \\
& \texttt{\textcolor{blue}{\{"error": "", "response": ""\}}} \\
& Continue this process until final result. Final "assistant" format should be: \\
& \texttt{\textcolor{blue}{Thought: ... Action: <API>Finish->answer</API> Answer: ...}} \\
& Or: \\
& \texttt{\textcolor{blue}{Thought: ...}} \\
&\texttt{\textcolor{blue}{Action: <API>Finish->give\_up\_and\_restart</API>}} \\ 
\midrule
Design Requirement & \small Design Q\&A around the \texttt{\{API\_NUM\}} APIs \texttt{\{API\_NAME\}}. Focus on collaboration between APIs, where later APIs utilize results from previous ones for further reasoning. \\ 
\bottomrule
\end{tabular}
\end{table}

This section details the creation of datasets used for training and evaluating \texttt{GraphChain}, including fine-tuning data and graph datasets across five real-world scenarios.

\subsection{Fine-tuning Dataset}

We constructed a comprehensive and robust dataset for fine-tuning \texttt{GraphChain} through a systematic, multi-faceted approach:

\begin{table}[t!]
\centering
\caption{\small Example of SFT data with step-by-step reasoning and API calls.}
\label{tab:example_of_SFT}
\begin{tabular}{|>{\columncolor{gray!10}\raggedright\arraybackslash}p{1.8cm}|>{\raggedright\arraybackslash}p{10.2cm}|}
\toprule
\rowcolor{gray!25}
\textbf{From} & \textbf{Value} \\ 
\midrule
User & \textit{How to identify the neighbor nodes of node 15 and analyze the fund flow between these neighbor nodes? Begin!} \\ 
\midrule
Assistant & \textbf{Thought:} To identify the neighbor nodes of node 15 and analyze fund flow between them, we need to proceed step by step. First, obtain all direct neighbor nodes of node 15 and transaction information. Then analyze fund flow between these neighbors. \\
& Let's obtain neighbor nodes and transaction information using \texttt{\textcolor{green}{<API>graph.out\_edges(15, data=True)</API>}} and \texttt{\textcolor{green}{<API>graph.in\_edges(15, data=True)</API>}} to get all outgoing and incoming edge information. \\
& \textbf{Action:} \texttt{\textcolor{green}{<API>graph.out\_edges(15, data=True)</API>}} \\ 
\midrule
Function & \texttt{\textcolor{blue}{\{"error": "", "response": "Direct outgoing transaction information of node 15: }}\\ 
& \texttt{\textcolor{blue}{- (15, 16, \{'weight': 400.0, 'date': '2022-06-10'\}) }}\\ 
& \texttt{\textcolor{blue}{- (15, 17, \{'weight': 200.0, 'date': '2022-07-01'\})"\}}} \\ 
\midrule
Assistant & \textbf{Thought:} Continue by obtaining incoming transaction information. \\
& \textbf{Action:} \texttt{\textcolor{green}{<API>graph.in\_edges(15, data=True)</API>}} \\ 
\midrule
Function & \texttt{\textcolor{blue}{\{"error": "", "response": "Direct incoming information for node 15:}} \\ 
&\texttt{\textcolor{blue}{- (10, 15, \{'weight': 880.0, 'date': '2022-05-20'\})"\}}} \\ 
\midrule
Assistant & \textbf{Thought:} Now we've identified neighbor nodes of 15 are 10, 16, and 17. Next, analyze fund flow between these neighbors using \texttt{\textcolor{green}{<API>graph.subgraph([10, 16, 17]).edges(data=True)</API>}}. \\
& \textbf{Action:} \texttt{\textcolor{green}{<API>graph.subgraph([10, 16, 17]).edges(data=True)</API>}} \\ 
\midrule
Function & \texttt{\textcolor{blue}{\{"error": "", "response": "(10, 16, \{'weight': 300.0, 'date': '2022-06-05'\}) }}\\ 
& \texttt{\textcolor{blue}{- (10, 17, \{'weight': 100.0, 'date': '2022-06-15'\})}} \\ 
& \texttt{\textcolor{blue}{- (16, 17, \{'weight': 50.0, 'date': '2022-07-10'\})"\}}} \\ 
\midrule
Assistant & \textbf{Thought:} Based on this information, we can draw conclusions: Node 10 transferred 880 yuan to node 15, 300 yuan to node 16, and 100 yuan to node 17. There is also fund flow between nodes 16 and 17, with node 16 transferring 50 yuan to node 17. \\
& \textbf{Action:} \texttt{\textcolor{green}{<API>Finish->answer</API>}} \\
& \textbf{Answer:} Node 10 not only transferred 880 yuan to node 15, but also transferred 300 yuan to node 16 and 100 yuan to node 17. There is also fund flow between nodes 16 and 17, specifically, node 16 transferred 50 yuan to node 17. \\ 
\bottomrule
\end{tabular}
\end{table}
\subsubsection{SFT Dataset Construction}


We curated 45 commonly used APIs from the NetworkX library based on relevance and usage frequency in graph-related tasks. To ensure diverse instruction coverage, we employed \texttt{ChatGPT} to generate various instructions tailored to these APIs. For each iteration, we randomly sampled APIs and prompted \texttt{ChatGPT} to reverse-engineer instructions centered around them, ensuring comprehensive coverage across the API set.

To enhance practical relevance, human experts crafted three exemplar instructions for each subgroup within five distinct real-world graph scenarios. These expertly designed prompts served as high-quality references, grounding the dataset in realistic use cases.

Our structured prompting strategy guided \texttt{ChatGPT} to produce outputs in a standardized format:
\begin{equation*}
\texttt{\{Thought: \ldots Action: \ldots\}}
\end{equation*}

Each action explicitly invoked an API with required parameters (e.g., \texttt{G.get\_edge\_data(8, 0, default=None)}). The outputs were fed into a code generator to produce executable code, which was then executed to obtain results formatted as:
\begin{equation*}
\texttt{\{"error": "\ldots", "response": "\ldots"\}}
\end{equation*}

These results were appended to the input for subsequent steps, creating a coherent action sequence. We introduced two auxiliary functions: \texttt{Finish->answer} (signaling successful task completion) and \texttt{Finish->giveup\_and\_restart} (allowing model reset and retry in cases of persistent errors).

Through this pipeline, we generated 9,986 (instruction, solution path) pairs that encapsulate a wide range of API-driven tasks reflecting the complexity of real-world graph-based problem-solving. Table~\ref{tab:example_of_SFT} shows an example from our SFT dataset.

\subsubsection{RL Dataset Construction}

\begin{table}[t!]
\centering
\caption{\small The prompt template for constructing the RL dataset.}
\label{tab:prompt_for_RL}
\begin{tabular}{|>{\columncolor{gray!10}}p{2cm}|p{10cm}|}
\toprule
\rowcolor{gray!25}
\textbf{Category} & \textbf{Description} \\ 
\midrule
Dataset Name & \texttt{cash\_flow\_graph.gexf} \\ 
\midrule
Dataset Type & \texttt{MultiDirected graph with weights and dates} \\ 
\midrule
Dataset Content & \small The fund transfer data of a specific group. Directed edge A→B means A transferred funds to B. Graph construction: \\
& \texttt{\textcolor{blue}{G = nx.MultiDiGraph(), G.add\_edge(sender, receiver, weight=amount, date=transfer\_date)}}, \\
& where "sender" and "receiver" are the transfer participants, "amount" is the money amount, and "transfer\_date" is the date. Integer type is used for nodes. \\ 
\midrule
Task & \small Judge the reasonableness of thought and API names based on three dimensions: \\
& \textbf{(1)} \textit{API Correctness}: Whether the method exists in \texttt{networkX}, accepts the specified parameters, and matches the dataset type. \\
& \textbf{(2)} \textit{Thought and API Effectiveness}: How directly and effectively this step contributes to solving the user question. \\
& \textbf{(3)} \textit{Graph Distillation}: Whether the current thought and API can reduce information content or narrow search scope. \\ 
\midrule
Output Type & \texttt{json} \\ 
\midrule
Output Rules & \small \textbf{(1)} Output must be json data containing "apiResult" as shown in the example. \\
& \textbf{(2)} Output can have only this json data with no additional information. \\ 
\midrule
Special Attention & \small Output in "apiResult" should follow: \\
& \texttt{\textcolor{blue}{\{"api\_Correctness": "", "thoughtAndApi\_Effectiveness": "", "Graph\_Distillation": ""\}}} \\ 
\bottomrule
\end{tabular}
\end{table}

For the reinforcement learning phase, we constructed a dataset with reward values for each step. We used GPT-4 to score each step based on three dimensions:
\begin{itemize}[leftmargin=*]
    \item \textbf{API Correctness}: Whether the tool invocation in the current step is valid
    \item \textbf{Thought and API Effectiveness}: The relevance of the tool selection to solving the query
    \item \textbf{Graph Distillation}: Whether the tool reduces the information content of the graph data
\end{itemize}

Table~\ref{tab:prompt_for_RL} shows the prompt template used for RL data construction.

Table~\ref{tab:examples_of_RL} shows two examples from our RL dataset with corresponding evaluation scores.

\begin{table}[h]
\centering
\caption{\small Examples from the RL dataset with evaluation scores.}
\label{tab:examples_of_RL}
\begin{tabular}{|>{\columncolor{gray!10}\raggedright\arraybackslash}p{1.8cm}|>{\raggedright\arraybackslash}p{4.2cm}|>{\raggedright\arraybackslash}p{6cm}|}
\toprule
\rowcolor{gray!25}
\textbf{Category} & \textbf{User Question} & \textbf{Thought and API} \\ 
\midrule
Input & \small Please analyze whether there is a fund circulation flow in the graph, and identify all nodes involved in the circulation flow. Additionally, please calculate the maximum amount of funds involved in these circulation paths. Begin! & \small We obtained the minimum cut of the maximum fund flow from node 1 to node 3 and its corresponding maximum flow value X. This means that, after considering all possible flow paths, the maximum fund transfer amount on the path from node 1 to node 3 is X. Additionally, we obtained the two sets that constitute the minimum cut, which helps us understand which nodes are involved in this maximum fund transfer process. The next step is to list the names of all nodes involved in the circulation flow to display directly to the user. \\
& & \texttt{\textcolor{green}{API: G.nodes()}} \\ 
\midrule
Output & & \texttt{\textcolor{blue}{API Correctness: 0.8, Thought and API Effectiveness: 0.9, Graph Distillation: 0.0}} \\ 
\midrule
Input & \small Please analyze whether there is a fund circulation flow in the graph, and identify all nodes involved in the circulation flow. Additionally, please calculate the maximum amount of funds involved in these circulation paths. Begin! & \small To complete this task, we need to proceed step by step. First, we need to determine whether there is a fund circulation flow in the graph, which can be achieved by calling \texttt{\textcolor{green}{<API>nx.find\_cycle(G)</API>}} to return a list of cycle edges if a cycle exists. Next, we need to identify all nodes involved in the circulation flow. \texttt{\textcolor{green}{<API>nx.simple\_cycles(G)</API>}} can be used to find all simple cycles (i.e., cycles that do not repeat any node) in the graph. Finally, for each identified circulation path, we need to calculate its maximum fund amount, which can be done by traversing each edge in the cycle and accumulating their weights. \\
& & \texttt{\textcolor{green}{API: nx.find\_cycle(G)}} \\ 
\midrule
Output & & \texttt{\textcolor{blue}{API Correctness: 0.9, Thought and API Effectiveness: 0.95, Graph Distillation: 1.0}} \\ 
\bottomrule
\end{tabular}
\end{table}

\subsection{Graph Dataset}

We extracted subgraphs from existing real-world datasets to construct graph datasets for five scenarios, each corresponding to different graph types:

\begin{itemize}[leftmargin=*]
    \item \textbf{Financial Networks}: From the Elliptic dataset, containing Bitcoin transaction graphs
    \item \textbf{Chemical Molecules}: From the QM9 dataset, including molecular structure graphs where atoms are nodes and chemical bonds are edges
    \item \textbf{Social Networks}: From the Facebook and Twitter datasets
    \item \textbf{Citation Graphs}: From the Cora, CiteSeer, and PubMed datasets
    \item \textbf{Traffic Networks}: From the METR-LA dataset
\end{itemize}

\begin{figure}[h]
    \centering
    \includegraphics[page=1,width=1.0\textwidth]{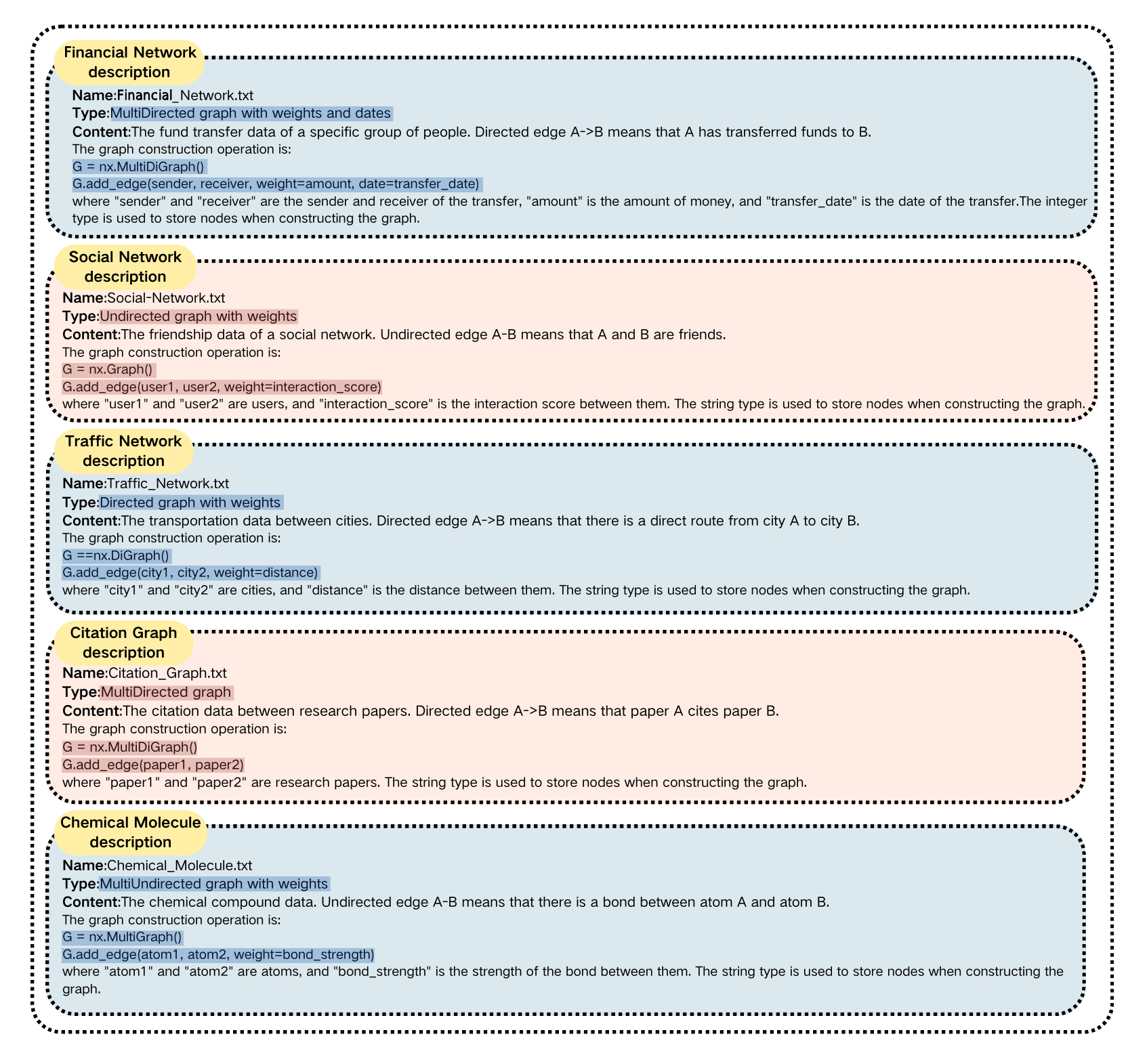}
    \caption{\small Detailed Description of the Graph Datasets for the Five Scenarios.}
    \label{fig:descrip_of_graph_data}
\end{figure}

For simplicity, we simplified the graph data as shown in Figure~\ref{fig:descrip_of_graph_data}. Following \citet{wang2024graphtool}, we prepared two versions of each graph to accommodate different baselines:

\begin{itemize}[leftmargin=*]
    \item For text-instruction baselines, we restricted inputs to no more than 30 nodes and 300 edges due to context length limitations
    \item For tool-instruction baselines, we limited inputs to no more than 100 nodes and 1000 edges
\end{itemize}

This approach ensures fair comparison across all methods while allowing us to evaluate \texttt{GraphChain}'s scalability advantages with full-sized graphs in our main experiments.

%% file: app/stta_complexity.tex
\section{Complexity Analysis of Structure-aware Test-Time Adaptation}
\label{app:stta_complexity}

This section analyzes the computational complexity of our Structure-aware Test-Time Adaptation (STTA) mechanism.

\subsection{Graph Structural Fingerprinting}

For a graph $G$ with $N$ nodes and $E$ edges, computing the $M$ smallest singular values of the normalized Laplacian has:
\begin{itemize}
    \item Time complexity: $\mathcal{O}(E \cdot M \cdot T_{\text{iter}})$, where $T_{\text{iter}}$ is the number of iterations in the iterative SVD algorithm
    \item Space complexity: $\mathcal{O}(N + E + NM)$
\end{itemize}

We employ iterative methods (Lanczos algorithm or power iterations) instead of full SVD to efficiently compute only the needed singular values. Since $M \ll N$ (typically $M = 10$ to $50$), this computation remains efficient even for large graphs.

\subsection{Structure-Conditioned Prompt Generation}

The adapter network $\mathcal{A}_{\psi}$ that maps the structural fingerprint to soft prompts has:
\begin{itemize}
    \item Time complexity: $\mathcal{O}(M \cdot H + H \cdot L_p \cdot d_{emb})$
    \item Space complexity: $\mathcal{O}(M \cdot H + H \cdot L_p \cdot d_{emb})$
\end{itemize}

Where $H$ is the hidden dimension, $L_p$ is the prompt length, and $d_{emb}$ is the embedding dimension. This adapter is extremely lightweight (0.01\%-0.1\% of LLM parameters).

\subsection{Self-Supervised Adaptation}

The REINFORCE-based adaptation using $K$ auxiliary queries with $R$ rollouts per query has:
\begin{itemize}
    \item Time complexity: $\mathcal{O}(K \cdot R \cdot \bar{N} \cdot C_{LLM} + K \cdot R \cdot \bar{N} \cdot C_{KL})$
    \item Space complexity: $\mathcal{O}(K \cdot R \cdot \bar{N} + |\psi|)$
\end{itemize}

Where $\bar{N}$ is the average chain length, $C_{LLM}$ is the cost of an LLM forward pass, $C_{KL}$ is the cost of computing KL divergence, and $|\psi|$ is the parameter count of the adapter.

\subsection{Overall Efficiency}

The total computational cost can be summarized as:
\begin{align}
C_{total} = \mathcal{O}(E \cdot M \cdot T_{iter}) + \mathcal{O}(K \cdot R \cdot \bar{N} \cdot C_{LLM}) + \mathcal{O}(T_{query} \cdot \bar{N}_{query} \cdot C_{LLM})
\end{align}

Our approach is efficient because: (1) graph fingerprinting is performed only once per graph; (2) adaptation requires few rollouts (typically $K=5$, $R=3$); and (3) only the small adapter network needs updating.

%% file: app/impact.tex
\newpage
\section{Broader Impact}
\label{app:impact}

GraphChain's ability to process large-scale graphs efficiently could significantly enhance data analysis capabilities in critical domains such as financial fraud detection, healthcare networks, and social network analysis. By enabling more effective reasoning over complex interconnected data, GraphChain could help identify suspicious transaction patterns, improve epidemiological network analysis, and better understand information propagation in social networks. The framework's adaptability across diverse graph structures makes it particularly valuable for interdisciplinary research and applications where domain experts need to analyze graph data without specialized technical knowledge. Moreover, the reduced computational requirements of our approach compared to retraining models for each new graph domain could lead to more environmentally sustainable AI deployments by decreasing the energy consumption associated with large-scale model training. These advancements contribute to more accessible, efficient, and effective graph analytics tools that can address various societal challenges.

%% file: neurips_2025.bbl
\begin{thebibliography}{46}
\providecommand{\natexlab}[1]{#1}
\providecommand{\url}[1]{\texttt{#1}}
\expandafter\ifx\csname urlstyle\endcsname\relax
  \providecommand{\doi}[1]{doi: #1}\else
  \providecommand{\doi}{doi: \begingroup \urlstyle{rm}\Url}\fi

\bibitem[Alfarra et~al.(2025)Alfarra, Correia, Ghanem, and Louizos]{alfarra2025test}
M.~Alfarra, A.~Correia, B.~Ghanem, and C.~Louizos.
\newblock Test-time adaptation with source based auxiliary tasks.
\newblock \emph{Transactions on Machine Learning Research}, 2025.

\bibitem[Anthropic(2024)]{TheC3}
A.~Anthropic.
\newblock The claude 3 model family: Opus, sonnet, haiku.
\newblock In \emph{Claude-3 Model Card}, 2024.

\bibitem[Beaudry and Renner(2011)]{beaudry2011intuitive}
N.~J. Beaudry and R.~Renner.
\newblock An intuitive proof of the data processing inequality.
\newblock \emph{arXiv preprint arXiv:1107.0740}, 2011.

\bibitem[Chai et~al.(2023)Chai, Zhang, Wu, Han, Hu, Huang, and Yang]{chai2023graphllm}
Z.~Chai, T.~Zhang, L.~Wu, K.~Han, X.~Hu, X.~Huang, and Y.~Yang.
\newblock Graphllm: Boosting graph reasoning ability of large language model.
\newblock \emph{arXiv preprint arXiv:2310.05845}, 2023.

\bibitem[Chen et~al.(2024{\natexlab{a}})Chen, Li, Tang, and Li]{DBLP:journals/corr/abs-2402-16029}
N.~Chen, Y.~Li, J.~Tang, and J.~Li.
\newblock Graphwiz: An instruction-following language model for graph problems.
\newblock \emph{CoRR}, abs/2402.16029, 2024{\natexlab{a}}.

\bibitem[Chen et~al.(2024{\natexlab{b}})Chen, Zhao, Jaiswal, Shah, and Wang]{chen2024llaga}
R.~Chen, T.~Zhao, A.~K. Jaiswal, N.~Shah, and Z.~Wang.
\newblock Llaga: Large language and graph assistant.
\newblock In \emph{International Conference on Machine Learning}, pages 7809--7823. PMLR, 2024{\natexlab{b}}.

\bibitem[Chen et~al.(2020)Chen, Chen, Xie, Cao, Gao, and Feng]{chen2020multi}
W.~Chen, L.~Chen, Y.~Xie, W.~Cao, Y.~Gao, and X.~Feng.
\newblock Multi-range attentive bicomponent graph convolutional network for traffic forecasting.
\newblock In \emph{Proceedings of the AAAI conference on artificial intelligence}, volume~34, pages 3529--3536, 2020.

\bibitem[Chen et~al.(2023)Chen, Zhou, Zhang, Gong, Zhao, and Wen]{chen2023chatcot}
Z.~Chen, K.~Zhou, B.~Zhang, Z.~Gong, W.~X. Zhao, and J.-R. Wen.
\newblock Chatcot: Tool-augmented chain-of-thought reasoning on chat-based large language models.
\newblock \emph{arXiv preprint arXiv:2305.14323}, 2023.

\bibitem[Du et~al.(2024)Du, Wei, and Zhang]{du2024anytool}
Y.~Du, F.~Wei, and H.~Zhang.
\newblock Anytool: Self-reflective, hierarchical agents for large-scale api calls.
\newblock In \emph{International Conference on Machine Learning}, pages 11812--11829. PMLR, 2024.

\bibitem[Fore et~al.(2024)Fore, Singh, and Stamoulis]{fore2024geckopt}
M.~Fore, S.~Singh, and D.~Stamoulis.
\newblock Geckopt: Llm system efficiency via intent-based tool selection.
\newblock In \emph{Proceedings of the Great Lakes Symposium on VLSI 2024}, pages 353--354, 2024.

\bibitem[Gao et~al.(2024)Gao, Shi, Zhu, Fang, Xin, Ren, Chen, Ma, and Ren]{gao2024confucius}
S.~Gao, Z.~Shi, M.~Zhu, B.~Fang, X.~Xin, P.~Ren, Z.~Chen, J.~Ma, and Z.~Ren.
\newblock Confucius: Iterative tool learning from introspection feedback by easy-to-difficult curriculum.
\newblock In \emph{Proceedings of the AAAI Conference on Artificial Intelligence}, volume~38, pages 18030--18038, 2024.

\bibitem[GLM(2024)]{glm2024chatglm}
T.~GLM.
\newblock Chatglm: A family of large language models from glm-130b to glm-4 all tools, 2024.

\bibitem[Gu et~al.(2024)Gu, Shu, Yu, Liu, Dong, Tang, Srinivasa, Latapie, and Su]{gu2024middleware}
Y.~Gu, Y.~Shu, H.~Yu, X.~Liu, Y.~Dong, J.~Tang, J.~Srinivasa, H.~Latapie, and Y.~Su.
\newblock Middleware for llms: Tools are instrumental for language agents in complex environments.
\newblock In \emph{Proceedings of the 2024 Conference on Empirical Methods in Natural Language Processing}, pages 7646--7663, 2024.

\bibitem[Guo et~al.(2025)Guo, Yang, Zhang, Song, Zhang, Xu, Zhu, Ma, Wang, Bi, et~al.]{guo2025deepseek}
D.~Guo, D.~Yang, H.~Zhang, J.~Song, R.~Zhang, R.~Xu, Q.~Zhu, S.~Ma, P.~Wang, X.~Bi, et~al.
\newblock Deepseek-r1: Incentivizing reasoning capability in llms via reinforcement learning.
\newblock \emph{arXiv preprint arXiv:2501.12948}, 2025.

\bibitem[Guo et~al.(2023)Guo, Du, Liu, Zhou, He, and Han]{guo2023gpt4graph}
J.~Guo, L.~Du, H.~Liu, M.~Zhou, X.~He, and S.~Han.
\newblock Gpt4graph: Can large language models understand graph structured data? an empirical evaluation and benchmarking.
\newblock \emph{arXiv preprint arXiv:2305.15066}, 2023.

\bibitem[Hu et~al.(2022)Hu, Shen, Wallis, Allen-Zhu, Li, Wang, Wang, Chen, et~al.]{hu2022lora}
E.~J. Hu, Y.~Shen, P.~Wallis, Z.~Allen-Zhu, Y.~Li, S.~Wang, L.~Wang, W.~Chen, et~al.
\newblock Lora: Low-rank adaptation of large language models.
\newblock \emph{ICLR}, 1\penalty0 (2):\penalty0 3, 2022.

\bibitem[Jaech et~al.(2024)Jaech, Kalai, Lerer, Richardson, El-Kishky, Low, Helyar, Madry, Beutel, Carney, et~al.]{jaech2024openai}
A.~Jaech, A.~Kalai, A.~Lerer, A.~Richardson, A.~El-Kishky, A.~Low, A.~Helyar, A.~Madry, A.~Beutel, A.~Carney, et~al.
\newblock Openai o1 system card.
\newblock \emph{arXiv preprint arXiv:2412.16720}, 2024.

\bibitem[Kulinski and Inouye(2023)]{kulinski2023towards}
S.~Kulinski and D.~I. Inouye.
\newblock Towards explaining distribution shifts.
\newblock In \emph{International Conference on Machine Learning}, pages 17931--17952. PMLR, 2023.

\bibitem[Leskovec and Mcauley(2012)]{leskovec2012learning}
J.~Leskovec and J.~Mcauley.
\newblock Learning to discover social circles in ego networks.
\newblock \emph{Advances in neural information processing systems}, 25, 2012.

\bibitem[Liang et~al.(2025)Liang, He, and Tan]{liang2025comprehensive}
J.~Liang, R.~He, and T.~Tan.
\newblock A comprehensive survey on test-time adaptation under distribution shifts.
\newblock \emph{International Journal of Computer Vision}, 133\penalty0 (1):\penalty0 31--64, 2025.

\bibitem[Liu et~al.(2024)Liu, Peng, Yi, Xie, Xiang, Liu, and Xu]{liu2024toolnet}
X.~Liu, Z.~Peng, X.~Yi, X.~Xie, L.~Xiang, Y.~Liu, and D.~Xu.
\newblock Toolnet: Connecting large language models with massive tools via tool graph.
\newblock \emph{arXiv preprint arXiv:2403.00839}, 2024.

\bibitem[Ma et~al.(2023)Ma, Zhang, Guo, and Xu]{ma2023swapprompt}
X.~Ma, J.~Zhang, S.~Guo, and W.~Xu.
\newblock Swapprompt: Test-time prompt adaptation for vision-language models.
\newblock \emph{Advances in Neural Information Processing Systems}, 36:\penalty0 65252--65264, 2023.

\bibitem[Mekala et~al.(2024)Mekala, Weston, Lanchantin, Raileanu, Lomeli, Shang, and Dwivedi-Yu]{mekala2024toolverifier}
D.~Mekala, J.~Weston, J.~Lanchantin, R.~Raileanu, M.~Lomeli, J.~Shang, and J.~Dwivedi-Yu.
\newblock Toolverifier: Generalization to new tools via self-verification.
\newblock In \emph{Findings of the Association for Computational Linguistics: EMNLP 2024}, pages 5026--5041, 2024.

\bibitem[Muhtar et~al.(2024)Muhtar, Shen, Yang, Liu, Lu, Liu, Zhan, Sun, Deng, Sun, et~al.]{muhtar2024streamadapter}
D.~Muhtar, Y.~Shen, Y.~Yang, X.~Liu, Y.~Lu, J.~Liu, Y.~Zhan, H.~Sun, W.~Deng, F.~Sun, et~al.
\newblock Streamadapter: Efficient test time adaptation from contextual streams.
\newblock \emph{arXiv preprint arXiv:2411.09289}, 2024.

\bibitem[OpenAI(2023)]{DBLP:journals/corr/abs-2303-08774}
OpenAI.
\newblock {GPT-4} technical report.
\newblock \emph{CoRR}, abs/2303.08774, 2023.

\bibitem[Qiao et~al.(2024)Qiao, Gui, Lv, Jia, Chen, and Zhang]{qiao2024making}
S.~Qiao, H.~Gui, C.~Lv, Q.~Jia, H.~Chen, and N.~Zhang.
\newblock Making language models better tool learners with execution feedback.
\newblock In \emph{Proceedings of the 2024 Conference of the North American Chapter of the Association for Computational Linguistics: Human Language Technologies (Volume 1: Long Papers)}, pages 3550--3568, 2024.

\bibitem[Qin et~al.(2023)Qin, Liang, Ye, Zhu, Yan, Lu, Lin, Cong, Tang, Qian, et~al.]{qin2023toolllm}
Y.~Qin, S.~Liang, Y.~Ye, K.~Zhu, L.~Yan, Y.~Lu, Y.~Lin, X.~Cong, X.~Tang, B.~Qian, et~al.
\newblock Toolllm: Facilitating large language models to master 16000+ real-world apis.
\newblock \emph{arXiv preprint arXiv:2307.16789}, 2023.

\bibitem[Shi et~al.(2024)Shi, Xu, Zhuang, Yu, Sun, Wu, Yang, and Wang]{shi2024medadapter}
W.~Shi, R.~Xu, Y.~Zhuang, Y.~Yu, H.~Sun, H.~Wu, C.~Yang, and M.~D. Wang.
\newblock Medadapter: Efficient test-time adaptation of large language models towards medical reasoning.
\newblock In \emph{Proceedings of the 2024 Conference on Empirical Methods in Natural Language Processing}, pages 22294--22314, 2024.

\bibitem[Shu et~al.(2022)Shu, Nie, Huang, Yu, Goldstein, Anandkumar, and Xiao]{shu2022test}
M.~Shu, W.~Nie, D.-A. Huang, Z.~Yu, T.~Goldstein, A.~Anandkumar, and C.~Xiao.
\newblock Test-time prompt tuning for zero-shot generalization in vision-language models.
\newblock \emph{Advances in Neural Information Processing Systems}, 35:\penalty0 14274--14289, 2022.

\bibitem[Su et~al.(2022)Su, Du, Yang, Zhou, Li, Rao, Sun, Lu, and Wen]{su2022molecular}
B.~Su, D.~Du, Z.~Yang, Y.~Zhou, J.~Li, A.~Rao, H.~Sun, Z.~Lu, and J.-R. Wen.
\newblock A molecular multimodal foundation model associating molecule graphs with natural language.
\newblock \emph{arXiv preprint arXiv:2209.05481}, 2022.

\bibitem[Suzgun et~al.(2025)Suzgun, Yuksekgonul, Bianchi, Jurafsky, and Zou]{suzgun2025dynamic}
M.~Suzgun, M.~Yuksekgonul, F.~Bianchi, D.~Jurafsky, and J.~Zou.
\newblock Dynamic cheatsheet: Test-time learning with adaptive memory.
\newblock \emph{arXiv preprint arXiv:2504.07952}, 2025.

\bibitem[Tang et~al.(2024)Tang, Yang, Wei, Shi, Su, Cheng, Yin, and Huang]{tang2024graphgpt}
J.~Tang, Y.~Yang, W.~Wei, L.~Shi, L.~Su, S.~Cheng, D.~Yin, and C.~Huang.
\newblock Graphgpt: Graph instruction tuning for large language models.
\newblock In \emph{Proceedings of the 47th International ACM SIGIR Conference on Research and Development in Information Retrieval}, pages 491--500, 2024.

\bibitem[Wang et~al.(2023{\natexlab{a}})Wang, Feng, He, Tan, Han, and Tsvetkov]{DBLP:conf/nips/WangFHTHT23}
H.~Wang, S.~Feng, T.~He, Z.~Tan, X.~Han, and Y.~Tsvetkov.
\newblock Can language models solve graph problems in natural language?
\newblock In A.~Oh, T.~Naumann, A.~Globerson, K.~Saenko, M.~Hardt, and S.~Levine, editors, \emph{Advances in Neural Information Processing Systems 36: Annual Conference on Neural Information Processing Systems 2023, NeurIPS 2023, New Orleans, LA, USA, December 10 - 16, 2023}, 2023{\natexlab{a}}.

\bibitem[Wang et~al.(2023{\natexlab{b}})Wang, Feng, He, Tan, Han, and Tsvetkov]{wang2023can}
H.~Wang, S.~Feng, T.~He, Z.~Tan, X.~Han, and Y.~Tsvetkov.
\newblock Can language models solve graph problems in natural language?
\newblock \emph{Advances in Neural Information Processing Systems}, 36:\penalty0 30840--30861, 2023{\natexlab{b}}.

\bibitem[Wang et~al.(2024{\natexlab{a}})Wang, Wu, Hou, Liu, Gao, and McAuley]{wang2024instructgraph}
J.~Wang, J.~Wu, Y.~Hou, Y.~Liu, M.~Gao, and J.~McAuley.
\newblock Instructgraph: Boosting large language models via graph-centric instruction tuning and preference alignment.
\newblock In \emph{Findings of the Association for Computational Linguistics ACL 2024}, pages 13492--13510, 2024{\natexlab{a}}.

\bibitem[Wang et~al.(2024{\natexlab{b}})Wang, Liang, Chen, Zhang, and Qin]{wang2024graphtool}
R.~Wang, S.~Liang, Q.~Chen, J.~Zhang, and K.~Qin.
\newblock Graphtool-instruction: Revolutionizing graph reasoning in llms through decomposed subtask instruction.
\newblock \emph{arXiv preprint arXiv:2412.12152}, 2024{\natexlab{b}}.

\bibitem[Wang et~al.(2025)Wang, Han, Ji, Wang, Baldwin, and Li]{DBLP:conf/iclr/WangHJWB025}
R.~Wang, X.~Han, L.~Ji, S.~Wang, T.~Baldwin, and H.~Li.
\newblock Toolgen: Unified tool retrieval and calling via generation.
\newblock In \emph{The Thirteenth International Conference on Learning Representations, {ICLR} 2025, Singapore, April 24-28, 2025}. OpenReview.net, 2025.

\bibitem[Weber et~al.(2019)Weber, Domeniconi, Chen, Weidele, Bellei, Robinson, and Leiserson]{weber2019anti}
M.~Weber, G.~Domeniconi, J.~Chen, D.~K.~I. Weidele, C.~Bellei, T.~Robinson, and C.~E. Leiserson.
\newblock Anti-money laundering in bitcoin: Experimenting with graph convolutional networks for financial forensics.
\newblock \emph{arXiv preprint arXiv:1908.02591}, 2019.

\bibitem[Wei et~al.(2022)Wei, Wang, Schuurmans, Bosma, Xia, Chi, Le, Zhou, et~al.]{wei2022chain}
J.~Wei, X.~Wang, D.~Schuurmans, M.~Bosma, F.~Xia, E.~Chi, Q.~V. Le, D.~Zhou, et~al.
\newblock Chain-of-thought prompting elicits reasoning in large language models.
\newblock \emph{Advances in neural information processing systems}, 35:\penalty0 24824--24837, 2022.

\bibitem[Wu et~al.(2018)Wu, Ramsundar, Feinberg, Gomes, Geniesse, Pappu, Leswing, and Pande]{wu2018moleculenet}
Z.~Wu, B.~Ramsundar, E.~N. Feinberg, J.~Gomes, C.~Geniesse, A.~S. Pappu, K.~Leswing, and V.~Pande.
\newblock Moleculenet: a benchmark for molecular machine learning.
\newblock \emph{Chemical science}, 9\penalty0 (2):\penalty0 513--530, 2018.

\bibitem[Xu et~al.(2023)Xu, Hong, Li, Hu, Chen, and Zhang]{xu2023tool}
Q.~Xu, F.~Hong, B.~Li, C.~Hu, Z.~Chen, and J.~Zhang.
\newblock On the tool manipulation capability of open-source large language models.
\newblock \emph{arXiv preprint arXiv:2305.16504}, 2023.

\bibitem[Yang et~al.(2016)Yang, Cohen, and Salakhudinov]{yang2016revisiting}
Z.~Yang, W.~Cohen, and R.~Salakhudinov.
\newblock Revisiting semi-supervised learning with graph embeddings.
\newblock In \emph{International conference on machine learning}, pages 40--48. PMLR, 2016.

\bibitem[Yao et~al.(2023)Yao, Zhao, Yu, Du, Shafran, Narasimhan, and Cao]{yao2023react}
S.~Yao, J.~Zhao, D.~Yu, N.~Du, I.~Shafran, K.~Narasimhan, and Y.~Cao.
\newblock React: Synergizing reasoning and acting in language models.
\newblock In \emph{International Conference on Learning Representations (ICLR)}, 2023.

\bibitem[Yu et~al.(2024)Yu, Wang, Ma, Guo, Zhan, Wang, Wu, Guo, and Zhang]{yu2024steptool}
Y.~Yu, Z.~Wang, W.~Ma, Z.~Guo, J.~Zhan, S.~Wang, C.~Wu, Z.~Guo, and M.~Zhang.
\newblock Steptool: A step-grained reinforcement learning framework for tool learning in llms.
\newblock \emph{arXiv preprint arXiv:2410.07745}, 2024.

\bibitem[Zhang(2023{\natexlab{a}})]{DBLP:journals/corr/abs-2304-11116}
J.~Zhang.
\newblock Graph-toolformer: To empower llms with graph reasoning ability via prompt augmented by chatgpt.
\newblock \emph{CoRR}, abs/2304.11116, 2023{\natexlab{a}}.

\bibitem[Zhang(2023{\natexlab{b}})]{zhang2023graph}
J.~Zhang.
\newblock Graph-toolformer: To empower llms with graph reasoning ability via prompt augmented by chatgpt.
\newblock \emph{arXiv preprint arXiv:2304.11116}, 2023{\natexlab{b}}.

\end{thebibliography}
